\documentclass[a4paper,11pt]{article}

\usepackage{amsmath,amssymb}
\usepackage{bm}
\usepackage{upgreek}
\usepackage{textcomp}
\usepackage{url}
\usepackage{rotating}
\usepackage{gensymb}
\usepackage{natbib}
\usepackage[english]{babel}
\usepackage[T1]{fontenc}
\usepackage{float}
\usepackage[top=1in, bottom=1.25in, left=1.25in, right=1.25in]{geometry}

\def\acknowledgments{\vskip12pt\noindent{\bf
Acknowledgments\vrule depth 6pt
width0pt\relax}\\*\noindent\ignorespaces}

\title{Inversion using a new low-dimensional representation of complex binary geological media based on a deep neural network}

\author{Eric Laloy\thanks{Belgian Nuclear Research Centre, Engineered and Geosystems Analysis. Email: {\tt elaloy@sckcen.be}}, Romain H\'erault\thanks{Normandie Univ, UNIROUEN, UNIHAVRE, INSA Rouen, LITIS, 76000 Rouen, France}, John Lee\thanks{IREC/MIRO and ICTEAM/MLG, Universit\'e catholique de Louvain}, Diederik Jacques\thanks{Belgian Nuclear Research Centre, Engineered and Geosystems Analysis}, and Niklas Linde\thanks{Applied and Environmental Geophysics Group, Institute of Earth Sciences, University of Lausanne}}
\date{September 30, 2017}
\begin{document}

\maketitle

\begin{abstract}
Efficient and high-fidelity prior sampling and inversion for complex geological media is still a largely unsolved challenge. Here, we use a deep neural network of the variational autoencoder type to construct a parametric low-dimensional base model parameterization of complex binary geological media. For inversion purposes, it has the attractive feature that random draws from an uncorrelated standard normal distribution yield model realizations with spatial characteristics that are in agreement with the training set. In comparison with the most commonly used parametric representations in probabilistic inversion, we find that our dimensionality reduction (DR) approach outperforms principle component analysis (PCA), optimization-PCA (OPCA) and discrete cosine transform (DCT) DR techniques for unconditional geostatistical simulation of a channelized prior model. For the considered examples, important compression ratios (200 - 500) are achieved. Given that the construction of our parameterization requires a training set of several tens of thousands of prior model realizations, our DR approach is more suited for probabilistic (or deterministic) inversion than for unconditional (or point-conditioned) geostatistical simulation. Probabilistic inversions of 2D steady-state and 3D transient hydraulic tomography data are used to demonstrate the DR-based inversion. For the 2D case study, the performance is superior compared to current state-of-the-art multiple-point statistics inversion by sequential geostatistical resampling (SGR). Inversion results for the 3D application are also encouraging.
\end{abstract}

\section{Introduction}
\label{intro}

Inverse modeling plays a fundamental role in subsurface hydrology and many other Earth science disciplines. Basically, one iteratively proposes new model perturbations that are consistent with a given prior model until the resulting forward response agrees with a set of measured data up to a pre-specified level. The inversion outcome commonly consists of one or more 2D or 3D subsurface property field(s) describing, for instance, hydraulic conductivity or porosity. Due to conceptual and numerical errors in the forward problem formulation, (input and output data) measurement errors and insufficient information content in the measured data, the inverse problem may not admit a unique solution. Hence, the inversion process should ideally provide an ensemble of geologic model realizations that accurately span the range of possible geologic models that honor both the available data used in the inversion and prior information. 

The prior model is often conveniently specified by a two-point geostatistical model or variogram, by assuming the subsurface structure to be multi-Gaussian. Whenever this simplifying assumption does not hold \citep[e.g.,][]{GomezHernandez-Wen1998, Journel-Zhang2006}, the prior model can instead be informed by  a so-called training image (TI). The TI is a large gridded 2D or 3D unconditional representation of the expected target spatial field that can be either continuous or categorical (e.g., geologic facies image). To generate model realizations that reproduce higher-order statistics found in the TI (together with possible direct point conditioning data), one can use a multiple-point statistics (MPS) simulation method \citep[e.g.,][]{Guardiano-Srivastava1993,Strebelle2002, Hu-Chugunova2008,Mariethoz2010b,Tahmasebi2012}. 

Since MPS does generally not rely on a explicit mathematical description of the prior model in terms of model parameters, it is impossible to use it for classical parameter-based inversion. This prohibits the use of gradient-based methods such as randomized maximum likelihood \citep[RML,][]{Kitanidis1995,Oliver1996} or regular Markov chain Monte Carlo (MCMC) sampling \citep{Robert-Casella2004,Brooks2011}. Note that this limitation also holds for the Markov mesh model \citep[MMM, ][]{Stien-Kolbjornsen2011} method despite the fact that it builds a statistical model describing transition probabilities of the facies distribution. Instead, one needs to resort to sampling from the prior model using MPS simulation. This process that is often referred to as sequential geostatistical resampling (SGR) inversion \citep{Hansen2008,Mariethoz2010a,Hansen2012} basically consists of iteratively generating new random perturbations from the prior model and to accept or reject them based on the resulting data misfit. The SGR inversion method is powerful in finding models that fit the data, but it suffers from two important drawbacks. As shown by \citet{Laloy2016}, (1) it tends to only explore the immediate vicinity of a single global minimum even when the objective function/posterior landscape is highly multi-modal and (2) may produce degraded geologic model realizations when fitting large datasets with high signal-to-noise ratios (SNR).

Although some recent work has demonstrated substantial improvements to SGR-based inversion \citep{Zahner2016}, in this study we take another direction and propose a new parametric dimensionality reduction approach for complex binary prior models. The driving idea is that if one can build a lower-dimensional model parameterization from which one can sample (after appropriate non-linear transformations) model realizations that are consistent with the TI, then global parameter-based inverse methods for moderately large continuous parameter spaces can be used. For instance, this opens up the possibility to explore the posterior model distribution using state-of-the-art adaptive MCMC sampling with DREAM$_{\rm \left(ZS\right)}$ \citep[][]{Vrugt2009,Laloy-Vrugt2012}. 

Previous work on using compressed parametric bases for hydrogeological inversion has relied on principal component analysis \citep[PCA, e.g.,][]{Reynolds1996,Sarma2006}, kernel-PCA \citep[e.g.,][]{Sarma2008}, level-set \citep[e.g.,][]{Dorn-Villegas2008}, discrete wavelet transforms \citep[e.g.,][]{Awotunde-Horne2013}, discrete cosine transform \citep[DCT, e.g.,][]{Jafarpour2010,Linde-Vrugt2013}, singular value decomposition \citep[SVD, e.g.,][]{Tavakoli-Reynolds2011}, and K-SVD \citep[e.g.,][]{Khaninezhad2012,Khaninezhad-Jafarpour2014} to name most of the strategies used. However, none of these bases define a manifold that is restricted to complex (non-Gaussian) geological models that are in strong agreement with a specific TI. In other words, randomly sampling the parameter space produces model realizations that are generally inconsistent with the TI. To the best of our knowledge, there is so far only the optimization-PCA approach (O-PCA) by \citet{Vo-Durlofsky2014,Vo-Durlofsky2015} that aims to build a lower-dimensional model representation that is fully consistent with the TI. The O-PCA essentially consists of a post-processing of a PCA model. To work it is necessary that the underlying PCA representation encodes sufficient information regarding important patterns in the TI model. As illustrated later on in this paper, this may not be the case when the TI contains highly connected features with a relatively large degree of variability. 

In this work, we use for the first time a deep neural network \citep[DNN, see, e.g.,][for an overview]{Goodfellow2016} to build a parametric low-dimensional representation of complex geologic models. More specifically, we train the generator of a deep variational autoencoder \citep[VAE,][]{Kingma-Welling2014} such that it generates TI-consistent geologic model realizations when fed with a low-dimensional standard Gaussian noise vector. A key characteristic of our VAE is that it is largely made up of convolutional layers \citep[see, e.g.,][]{Lecun1998, Krizhevsky2012}. This type of neural layer is based on an adaptive convolutional filter and is well suited to image processing \citep[][]{Krizhevsky2012,Goodfellow2016}. Our model generator is then used within a Bayesian inversion framework to sample the posterior distribution of binary 2D and 3D subsurface property fields. 

This paper is organized as follows. Section \ref{methods} presents the different elements of our dimensionality reduction (DR) approach and the selected inversion framework. This is followed in section \ref{results_dr} with the analysis of its performance in generating geostatistical realizations using 2D and 3D TIs with and without conditioning on direct point data (also referred to as hard data), together with some comparisons against other existing DR methods. Synthetic 2D and 3D experiments involving both steady-state and transient groundwater flow are then used to demonstrate our proposed DR-based inversion approach. In section \ref{discussion} we discuss the advantages and limitations of our method and outline possible future developments. Finally, section \ref{conclusion} concludes with a summary of the most important findings.

\section{Methods}
\label{methods}
\subsection{Deep neural network architecture}
\label{methods_architecture}

\subsubsection{Generalities}
\label{methods_architecture_gen}

We consider a deep neural network that belongs to the class of autoencoders \citep[AEs, see, e.g.,][]{Goodfellow2016}. Neural networks basically define the (possibly complex) relationships existing between input, $\textbf{x}$, and output, $\textbf{y}$, data vectors by using combinations of computational units that are called neurons. A neuron is an operator of the form:

\begin{equation}
h\left(\textbf{x}\right) =f\left(\langle \textbf{x}, \textbf{w} \rangle+ b \right),
\label{dnn1}
\end{equation}

where $h\left(\cdot \right)$ is the scalar output of the neuron, $f\left(\cdot \right)$ is a nonlinear function that is called the ``activation function", $\langle\cdot,\cdot\rangle$ signifies the scalar product, $\textbf{w} = \left[w_1, \cdots, w_N\right]$ is a set of weights of same dimension, $N$, as $\textbf{x}$ and $b$ represents the bias associated with the neuron. For a given task, the values for $\textbf{w}$ and $b$ associated with each neuron must be optimized or ``learned" such that the resulting neural network performs as well as possible. When $f\left(\cdot \right)$ is differentiable, $\textbf{w}$ and $b$ can be learned by gradient descent. Common forms of $f\left(\cdot \right)$ include the rectified linear unit (ReLU), sigmoid function and hyperbolic tangent function.

When there is no directed loops or cycles across neurons or combinations thereof, the network is said to be feedforward (FFN). In the FFN architecture, the neurons are organized in layers. A standard layer is given by

\begin{equation}
\textbf{h}\left(\textbf{x}\right)=f\left(\textbf{W}\textbf{x} + \textbf{b} \right),
\label{dnn2}
\end{equation}

where $\textbf{W}$ and $\textbf{b}$ are now a matrix of weights and a vector of biases, respectively. The name multilayer perceptron (MLP) designates a FFN with more than one layer. A most typical network is the 2-layer MLP, which consists of two layers with the outputs of the first-layer neurons becoming inputs to the second-layer neurons

\begin{equation}
\textbf{y}=\textbf{g}\left[\textbf{h}\left(\textbf{x}\right)\right]  \equiv f_2\left[\textbf{W}_2\cdot f_1\left(\textbf{W}_1\textbf{x} + \textbf{b}_1 \right) + \textbf{b}_2 \right],
\label{dnn3}
\end{equation}

where $\textbf{g}\left(\cdot\right)$ and $\textbf{h}\left(\cdot\right)$ are referred to as output layer and hidden layer, respectively.

In theory, the two-layer MLP described in equation (\ref{dnn3}) is a universal approximator as it can approximate any underlying process between $\textbf{y}$ and $\textbf{x}$ \citep{Cybenko1989,Hornik1991}. However, this only works if the dimension of $\textbf{h}\left(\cdot\right)$ is (potentially many orders of magnitudes) larger than that of the input $\textbf{x}$, thereby making learning practically infeasible and the two-layer MLP approximator useless for large $N$ (typically $N \geq$ 10-25). For high-dimensional input data such as images, researchers have found that it is much more efficient to use many hidden layers rather than increasing the size of a single hidden layer \citep[e.g.,][]{Goodfellow2016}. When a FFN/MLP has more than one hidden layer it is considered to be deep. Nevertheless, current deep networks are not necessarily purely FFN but may mix different aspects of FFN, such as convolutional neural networks (CNN, see section \ref{methods_architecture_cnn} below) and recurrent neural networks \citep[RNN, see, e.g.,][]{Goodfellow2016}.

\subsubsection{Variational autoencoders}
\label{methods_architecture_vae}

\begin{figure}[H]
	\noindent\hspace{-0.25cm}\includegraphics[width=35pc]{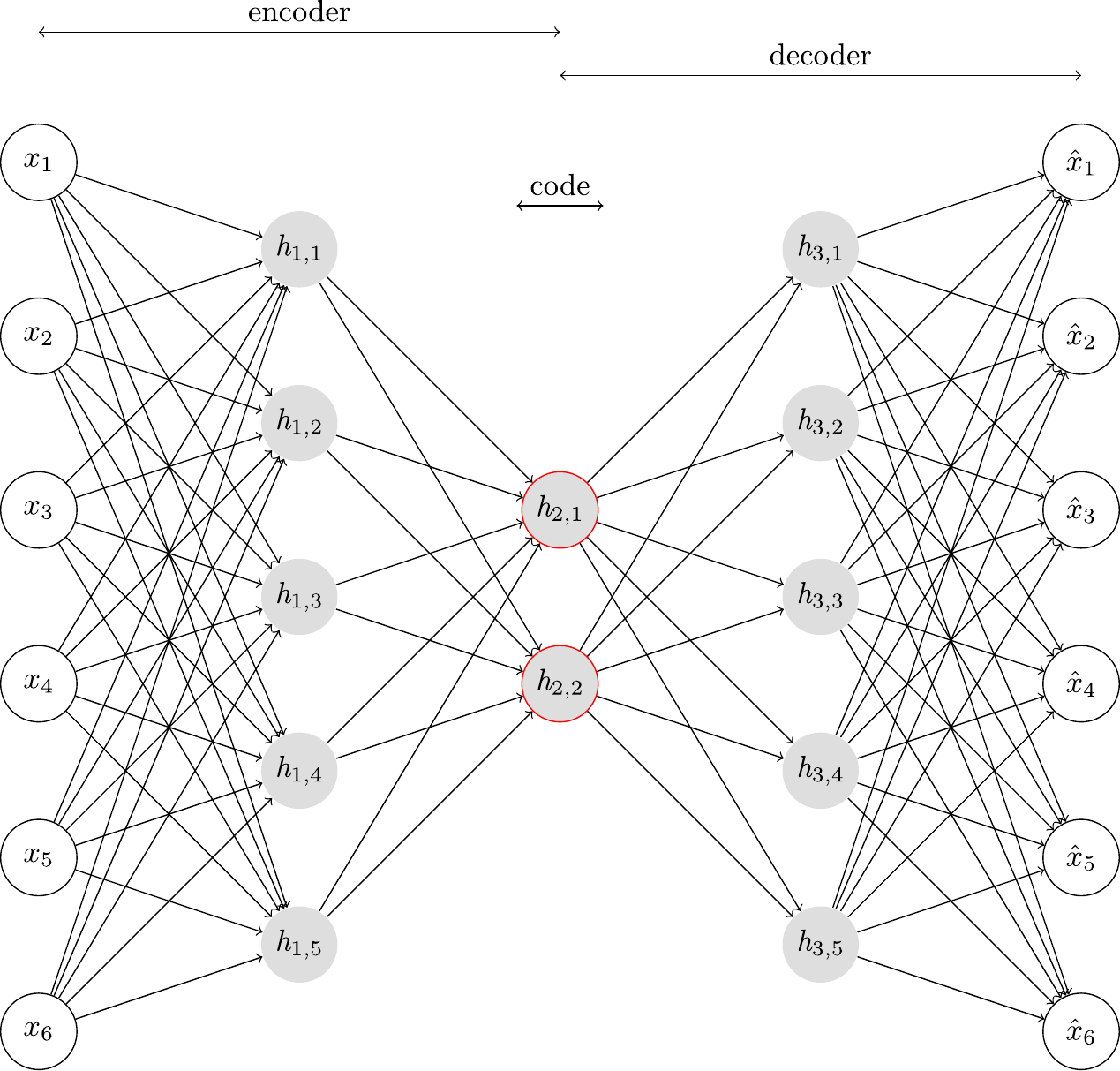}
	\caption{Schematic example of an autoencoder with three hidden layers, $\textbf{h}_1$, $\textbf{h}_2$ and $\textbf{h}_3$. The $\textbf{x} = \left[x_1, \cdots, x_6\right]$ vector is a 6-dimensional input feature, the $h_{\left(\cdot,\cdot\right)}$ denote the hidden layers with (red-contoured) $h_{\left(2,\cdot\right)}$ the central layer, or ``code", where the largest dimensionality reduction is performed and the $\hat{\textbf{x}} = \left[\hat{x}_1, \cdots, \hat{x}_6\right]$ vector signifies the reconstructed input.}
	\label{fig1}
\end{figure}

An AE is a deep neural network that defines a reversible, nonlinear low-dimensional parameterization of (higher dimensional) input data. Consequently, it has an hourglass-like shape in terms of neural layer sizes (see simplified representation in Figure \ref{fig1}). The most central layer is referred to as the ``code" and defines the low-dimensional space, while the output layer has the same dimensionality as the input layer. The part of the network that connects the input to the code is called the ``encoder", while the one that connects the code to the output is referred to as the ``decoder". 

Autoencoders are generative, which means that, after appropriate training (see section \ref{methods_training}), they can be used to generate new pattern realizations that are consistent with those found in a given set of (training) features. However, only the class of AEs formed by the so-called variational autoencoders (VAE) can use white noise as input to generate new patterns. In a geostatistical context, this implies that the VAE can be trained such that it randomly generates new geologic model realizations that honor the higher-order statistics found in a set of training images. How to achieve this is further described later on in this section and in section \ref{methods_training}. 

\begin{figure}[H]
	\noindent\hspace{-1cm}\includegraphics[width=40pc]{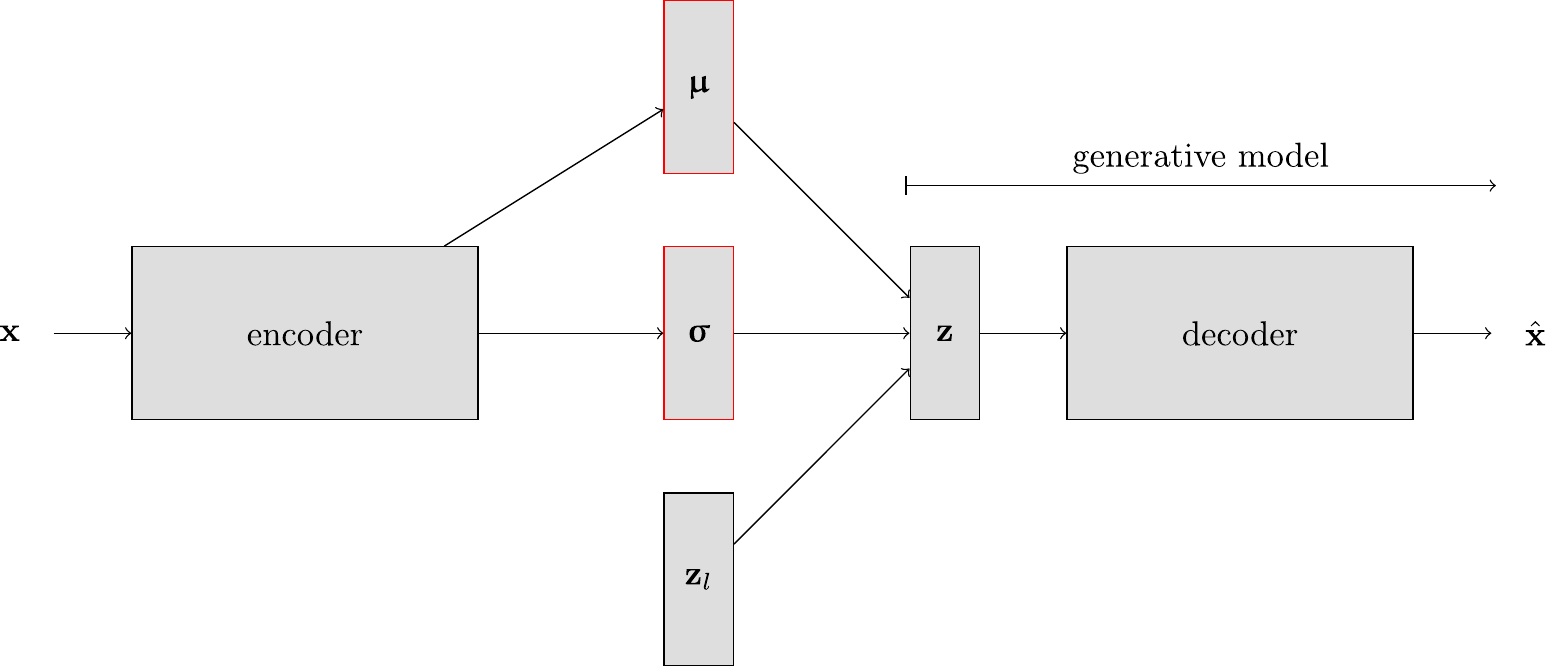}
	\caption{Illustration of a variational deep autoencoder (VAE). The boxes labeled with the $\bm{\upmu}$ and $\bm{\upsigma}$ letters represent the low-dimensional mean and standard deviation vectors, respectively, that are encoded from the input data. During training, $\textbf{z}_l$ is drawn from the fixed $p\left(\textbf{z}_l\right)$ distribution and then rescaled into $\textbf{z} = \textbf{z}_l\times\bm{\upsigma} + \bm{\upmu}$ (where $\times$ means element-wise multiplication). Training aims at jointly minimizing (1) the data reconstruction loss and the deviations between (2) $\bm{\upmu}$ and the zero-vector of the same size and (3) $\bm{\upsigma}$ and the unit vector of the same size. The ensemble of tasks (2) and (3) is called the ``reparameterization trick" \citep{Kingma-Welling2014}. For subsequent stochastic data generation, the process starts with sampling $\textbf{z}$ from the fixed $p\left(\textbf{z}\right)$ distribution.}
	\label{fig2}
\end{figure}

A VAE network can be summarized as follows. The code consists of two lower-dimensional vectors: a vector of means, $\bm{\upmu}$, and a vector of standard deviations, $\bm{\upsigma}$ (Figure \ref{fig2}). The first element of the decoder is a randomly sampled vector of ``latent" standard normal variables, $\textbf{z}_l$, of the same size, $d$, as $\bm{\upmu}$ and $\bm{\upsigma}$. The $\textbf{z}_l$ vector is subsequently rescaled into $\textbf{z} = \textbf{z}_l\times\bm{\upsigma} + \bm{\upmu}$ (where $\times$ means element-wise multiplication), and the so-produced $\textbf{z}$ vector continues its journey through the decoder to eventually produce the output, $\hat{\textbf{x}}$ (Figure \ref{fig2}). Here the symbol $\hat{\textbf{x}}$ is used instead of $\textbf{y}$ to make it clear that for this application the output is a reconstruction of the input, $\textbf{x}$. As detailed in section \ref{methods_training}, the $\textbf{w}$ vectors and $b$ values that are associated with every neuron of every layer of the network (encoder part, $\bm{\upmu}$, $\bm{\upsigma}$ and decoder part) are jointly optimized by gradient descent such that (1) the differences between $\textbf{x}$ and $\hat{\textbf{x}}$ are minimized and (2) the $d$-dimensional $\textbf{z}$ vector conforms as closely as possible to a $N\left(\textbf{0}_d,\textbf{I}_d\right)$ with $\textbf{0}_d$ a $d$-dimensional zero-vector and $\textbf{I}_d$ the $d \times d$ identity matrix. After training, new model realizations can be generated by sampling $\textbf{z} \propto N\left(\textbf{0}_d,\textbf{I}_d\right)$ and running the $\textbf{z}$ samples through the decoder. 

At this point, we would like to re-emphasize that it is the combination of (1) a low-dimensional parameterization in terms of a multivariate standard normal distribution and (2) the ability to map a given set of low-dimensional parameter values into complex MPS geostatistical model realizations that makes our presented method unique and suitable for MPS-based inversion.

\subsubsection{Convolutional layers}
\label{methods_architecture_cnn}

The last salient characteristic of our deep VAE is the use of convolutional layers at various levels of both the encoder and the decoder. The convolutional layer is the main building block of the CNN-type of architecture. It is particularly well suited for image processing applications because it explicitly accounts for the spatial structure of the input data, whether the input image, $\textbf{X}$, or the incoming hidden layer in the network. When the input is a 2D image (with possibly 3 channels for a RGB image), a convolutional layer, $\textbf{h}^k$, is built from a series of $k = 1, \cdots, N_k$ small $N_i \times N_j$ filters, $\textbf{w}^k$, that convolve an input pixel, $X_{m,n}$ to $h^{k}_{m,n}$ as

\begin{equation}
h^k_{m,n}\left(X_{m,n}\right) = f\left(\sum_{i = 1,j=1}^{N_i,N_j} w^{k}_{i,j}X_{m+i,n+j} + b_k \right),
\label{dnn4}
\end{equation}

where for computational efficiency $f\left(\cdot \right)$ is typically a ReLU: $f\left(x \right) = \max\left(0,x\right)$. The so-produced ensemble of $N_k$ $\textbf{h}^k$ ``feature maps" forms a volume called the convolutional representation. The larger the $N_k$, the potentially more comprehensive is the representation of the input data. Other important convolution parameters that we do not discuss here are the ``stride", that is, the degree of overlapping between successive moves in the forward pass of a given filter, and the ``zero-padding" which involves padding the borders of the input image or volume for size preservation. For further information, we refer the reader to \citet{Goodfellow2016} and online tutorials\footnote{For instance: \url{http://deeplearning.net/tutorial/} and \url{http://cs231n.github.io/convolutional-networks/}.}. 

\begin{figure}[h!]
	\noindent\hspace{0cm}\includegraphics[width=30pc]{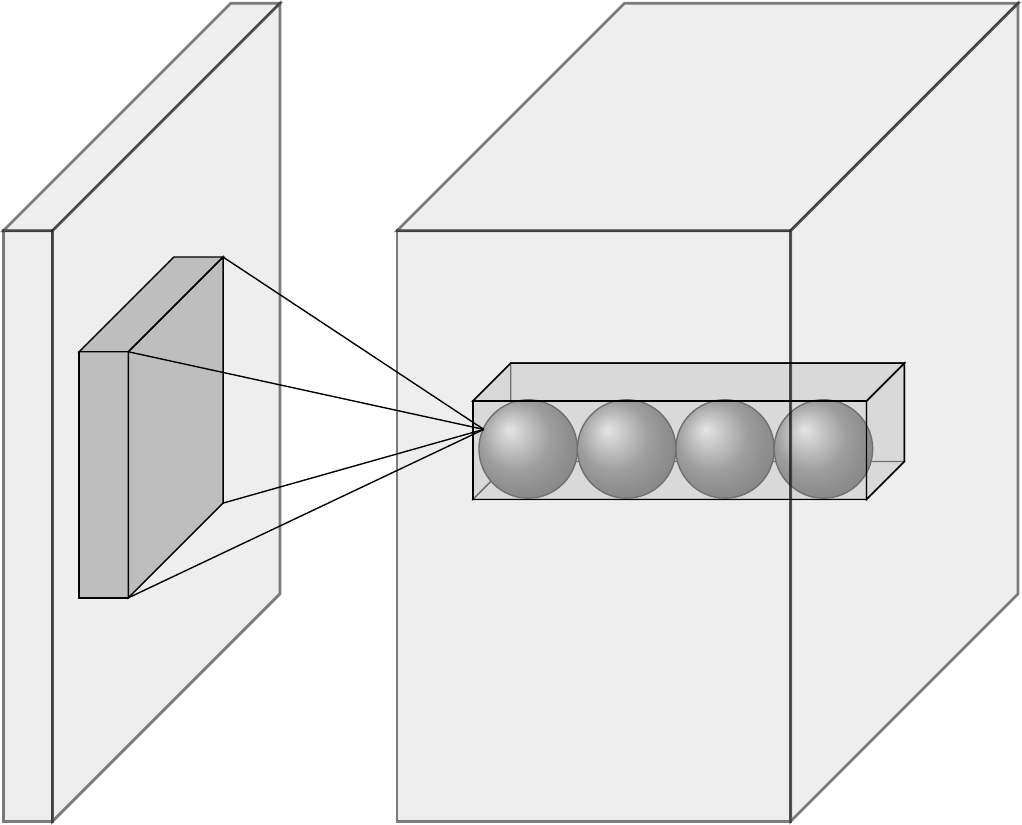}
	\caption{Stylized representation of convolutional layer neurons connected to their receptive field (picture inspired by \protect\url{http://cs231n.github.io/convolutional-networks/}).}
	\label{fig3}
\end{figure}

Note that each neuron of a given convolutional layer (volume) is connected to only a local region of the input image or volume, called ``receptive field" (see Figure \ref{fig3}). This enforces spatially-local exploration at the level of one convolutional layer. The encoded spatial information includes increasingly larger patterns when the number of stacked convolutional layers increases. Also, recall that the elements (pixels/voxels) of a given feature map share the same $\textbf{w}^k$ and $b_k$. The number of CNN parameters to be optimized is thus greatly reduced compared to that of a standard fully-connected (FC) (or dense) FFN of similar depth. Besides the convolutional layer itself, the other important component of a CNN is the ``pooling" layer. Pooling is a form of nonlinear down-sampling used to reduce the dimensions of a convolutional layer and thereby limit overfitting. It proceeds by converting small sub-regions of equal size into single values, such as the maximum or the mean. Using the maximum (``max-pooling") is the most common practice. When used, pooling always follows a series of one or more convolution operations.

\subsubsection{Practical implementation}
\label{methods_architecture_details}

We now have described all of the main ingredients of our devised deep VAECNN network, which is depicted in Figure 4. For the sake of brevity, the various individual fully-connected (FC), variational (Q), convolutional (C) and pooling (P) layers (see Figure 4) are not detailed in this paper but the corresponding computer code is available from the first author. Our VAECNN was implemented within the open-source LASAGNE Python software \citep{lasagne} which works on top of the open-source THEANO Python library \citep{theano}. In addition, note that for each result presented in this study, the $\textbf{z}$ vector is always 50-dimensional no matter the dimensions of the considered geologic model domain (from $100 \times 100$ to $30 \times 32 \times 27$). 

Unless stated otherwise, model generation is performed by recycling the initially produced model (bottom blue architecture in Figure 4) through the full network (top light red architecture in Figure 4) for 10 times. This sequential relooping generally improves the quality of the generated model realizations. The very last step of our generation process is an hard thresholding of the produced model. We used the 0.5 cutoff value although the classical Otsu's method could be used as well \citep[][]{Otsu1979}. Fortunately, the impact of relooping on generation time remains quite acceptable. For the models considered herein, generating a realization without relooping takes about 0.01 s - 0.03 s on a modern workstation. Including the 10 successive cycles then incurs a total generation time in the range 0.2 s - 0.4 s.

\begin{sidewaysfigure}[H]
	\noindent\hspace{0cm}\includegraphics[width=50pc]{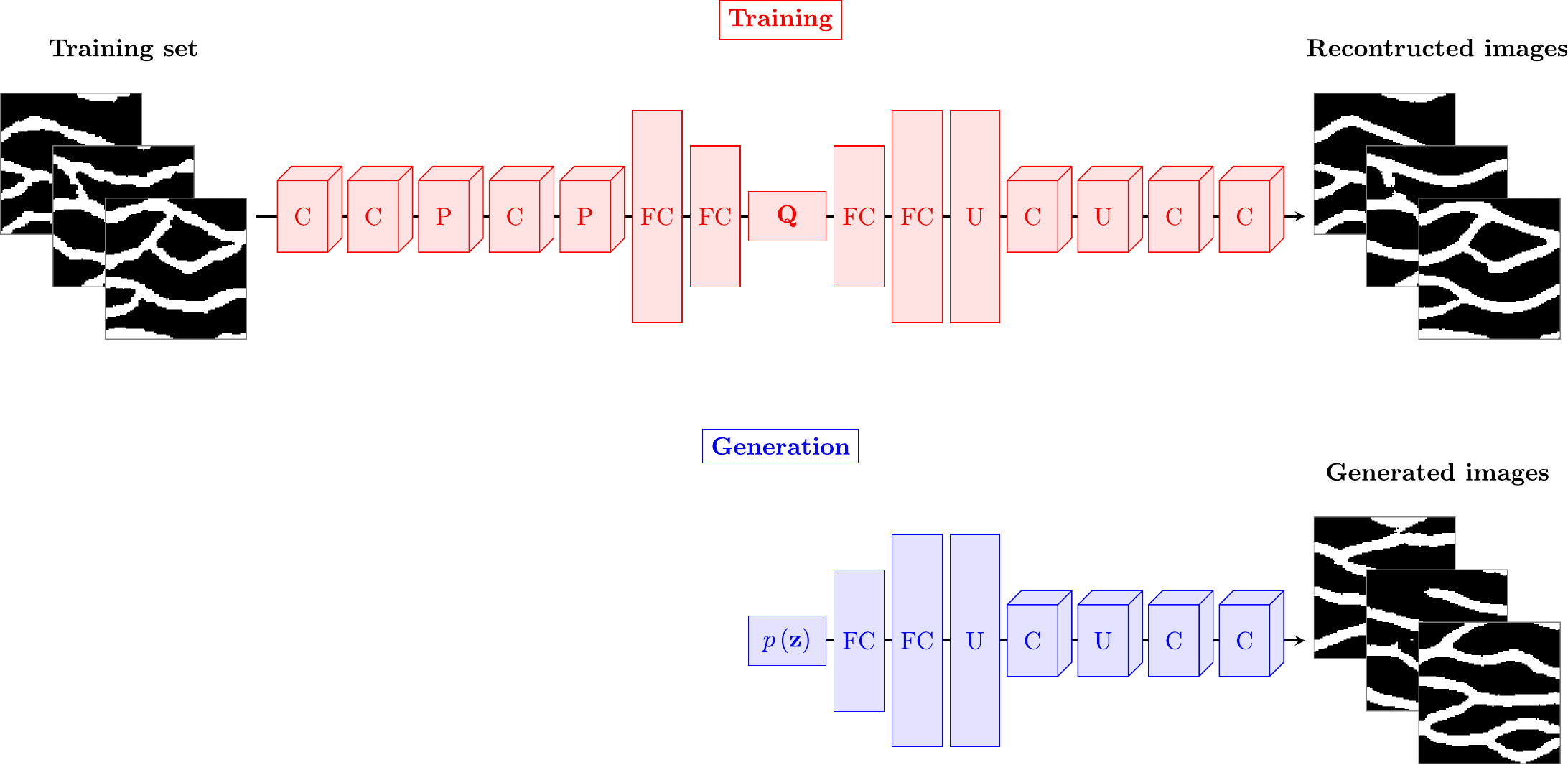}
	\caption{Overview of the VAECNN used herein. The letters C, P, FC and U stand for convolutional layer, pooling layer, fully connected layer and upscaling, respectively. In the upper (red) architecture (used for training), the box labeled with the Q letter includes the $\bm{\upmu}$, $\bm{\upsigma}$, $\textbf{z}_l$ and $\textbf{z}$ labeled boxes of Figure \ref{fig2}. The process of stochastic model generation (bottom blue architecture) starts with sampling the low-dimensional vector $\textbf{z}$ from its prespecified distribution $p\left(\textbf{z}\right)$. In this study, $\textbf{z}$ is always 50-dimensional.}
	\label{fig4}
\end{sidewaysfigure}

\subsection{Training the deep variational autoencoder}
\label{methods_training}

Considering the case where a $100 \times 100$ binary image is generated from a 50-dimensional $\textbf{z}$, the network described in Figure 4 includes 34,039,237 optimized parameters (weights and bias associated with neurons of dense layers and filters of convolutional layers). Such a high number of optimized parameters is not unusual for applications of deep neural networks. As the activation functions are differentiable, using an objective function that is also differentiable makes it possible to optimize/train the network by stochastic gradient descent (that is, gradient descent using a series of mini-batches rather than all the data at once) together with back propagation. This means that the objective function derivative is propagated backwards throughout the network using the chain rule, in order to update the parameters. Various stochastic gradient descent algorithms are available. In this work, we used the adaptive moment estimation (ADAM) algorithm which has been proven efficient for different types of deep networks \citep{Kingma-Ba2015}.

The objective or loss function, $\mathcal{L}$, must be set such that at training time: (1) the differences between reconstructed images, $\hat{\textbf{X}}$, and their original counterparts, $\textbf{X}$, are minimized and (2) the $d$-dimensional random $\textbf{z}$ vector conforms as closely as possible to a $N\left(\textbf{0}_d,\textbf{I}_d\right)$. This translates into the following formulation \citep[][]{Kingma-Welling2014,Gregor2015}

\begin{equation}
\mathcal{L} = \mathcal{L}^{x} + \mathcal{L}^{z},
\label{dnn5}
\end{equation}

with

\begin{equation}
\mathcal{L}^{x} = -\log\left[p\left(\textbf{X} | \textbf{z}\right)\right], 
\label{dnn6}
\end{equation}

\begin{equation}
\mathcal{L}^{z} = D_{KL}\left[q\left(\textbf{z}|\textbf{X}\right)||p\left(\textbf{z}\right)\right],
\label{dnn7}
\end{equation}

where $p\left(\textbf{z}\right)$ is the probability density function (pdf) of the code $\textbf{z}$ (here $p\left(\textbf{z}\right)\equiv N\left(\textbf{0}_d,\textbf{I}_d\right)$) and the conditional distribution $ p\left(\textbf{X} | \textbf{z}\right)$ is referred to as a probabilistic decoder, since given $\textbf{z}$ it generates a probability distribution over the possible corresponding values of $\textbf{X}$. Similarly, $q\left(\textbf{z}|\textbf{X}\right)$ denotes a probabilistic encoder, as given a feature $\textbf{X}$ it creates a distribution over the possible values of $\textbf{z}$ from which $\textbf{X}$ could have been produced \citep{Kingma-Welling2014}. To ensure that $q\left(\textbf{z}|\textbf{X}\right)$ is close to the targeted $p\left(\textbf{z}\right)$, the Kullback-Leibler divergence ($D_{KL}$) from $q\left(\textbf{z}|\textbf{X}\right)$ to $p\left(\textbf{z}\right)$, $D_{KL}\left[q\left(\textbf{z}|\textbf{X}\right)||p\left(\textbf{z}\right)\right]$ is minimized (equation (\ref{dnn7})).

If $\textbf{X}$ is a binary image, equation (\ref{dnn6}) can be rewritten as a binary cross-entropy minimization

\begin{equation}
\mathcal{L}^{x} =\sum\limits_{i=1}^{N}\left\{-X_i\log\left[\hat{X}_i\right]-\left(1-X_i\right)\log\left[\left(1-\hat{X}_i\right)\right]\right\}.
\label{dnn8}
\end{equation}

In addition, for $p\left(\textbf{z}\right)\equiv N\left(\textbf{0}_d,\textbf{I}_d\right)$ equation \ref{dnn5} becomes

\begin{equation}
\mathcal{L}^{z} = \displaystyle\frac{1}{2}\left[\sum\limits_{i=1}^{d}\left(\mu_i^2+\sigma_i^2-\log\sigma_i^2\right) \right]-\displaystyle\frac{d}{2},
\label{dnn9}
\end{equation}

where the $\mu_i$ and $\sigma_i$ correspond to the i$^{th}$ elements of the $\bm{\upmu}$ and $\bm{\upsigma}$ vectors displayed in Figure \ref{fig2}, respectively. Working with $\bm{\upmu}$ and $\bm{\upsigma}$ vectors rather than with the $\textbf{z}$ vector directly is called the ``reparameterization trick" \citep[for details, see][]{Kingma-Welling2014}.

Finally, in practice one may need to weight the components of $\mathcal{L}$, leading to

\begin{equation}
\mathcal{L} = \mathcal{L}^{x} + \alpha\mathcal{L}^{z},
\label{dnn10}
\end{equation}

where $\alpha$ is a weight factor that we set after limited trial and error either to 20 or to 40 depending on the application.

The minimization of equation (\ref{dnn10}) was performed on a GPU Tesla K40 and training the VAECNN for 100 epochs (full cycles of the stochastic gradient descent) took between 5 and 13 hours, depending on the size of the model domain (from $100 \times 100$ to $32 \times 30 \times 27$), the number of training images (from 19,000 to 80,000), and the exact size of the network. Computing times will be further detailed below for each test case individually.

\subsection{Assessing geostatistical simulation quality}

Although inversion is our primary objective, assessing the quality of the geostatistial realizations produced by our DR approach is important as it will determine the ultimate quality of our inversion results. Following \citet{Tan2014}, we considered both how well the patterns in the training images are reproduced by the DR-based realizations and the between-realization variability, also termed ``space of uncertainty". 

Besides visual inspection, we estimated closeness with the training images by means of the two-point cluster function \citep[CF,][]{Torquato1988} which is also called connectivity function \citep{Pardo-Dowd2003}. The CF is the probability that there exists a continuous path of the same facies between two points of the same facies separated by a given lag distance \citep[see,][for mathematical details]{Torquato1988,Pardo-Dowd2003}. Using the implementation by \citet{Lemmens2017}, the CF was calculated for each facies along the $x$, $y$, and main diagonal, $d_{xy}$, directions for the 2D case studies. For the 3D case studies, we also considered the $z$ and main diagonal $d_{xz}$ and $d_{yz}$ directions. This resulted into a total of six CF curves for each combination of indicator and facies.

The so-called space of uncertainty is a measure of the between-realizations variability associated with given geostatistical simulation algorithm. We characterized the space of uncertainty for the 2D case using an average distance between multiple-point histograms (MPH) over 100 realizations \citep[see][for details]{Tan2014}. This was done both for the training set and DR-based realizations. The used distance is the so-called Jensen-Shannon (JS) divergence \citep{Cover-Thomas1991,Tan2014}.

\begin{equation}
d_{JS}\left(\textbf{m}^{k},\textbf{m}^{k^\prime},\right) = \frac{1}{2}\sum\limits_{i=1}^{N_b}MPH_i^{k}\log\left[\frac{MPH_i^{k}}{MPH_i^{k^\prime}}\right] + \frac{1}{2}\sum\limits_{i=1}^{N_b}MPH_i^{k^\prime}\log\left[\frac{MPH_i^{k^\prime}}{MPH_i^{k}}\right],
\label{spunc0}
\end{equation}

where $\textbf{m}^{k}$ and $\textbf{m}^{k^\prime}$ are two different model realizations and $MPH_1^k, \cdots, MPH_{Nb}^k$ are the components of the $\textbf{MPH}^{k}$ vector calculated for realization $\textbf{m}^{k}$. Since the MPH was computed for a $4 \times 4$ template, $Nb$ was $2^{16} = 65,536$.

The space of uncertainty, defined here as the average distance, $\overline{d_{JS}}$, for the set of either $K$ training images or $K$ DR-based realizations, is given by

\begin{equation}
\overline{d_{JS}} = \frac{1}{K\left(K-1\right)} \sum\limits_{k=1}^{K}\sum\limits_{k^\prime=1}^{K} d_{JS}\left(\textbf{m}^{k},\textbf{m}^{k^\prime},\right).
\label{spunc1}
\end{equation}

\subsection{Bayesian inversion}
\label{methods_bayes}

A common representation of the forward problem is

\begin{equation}
\textbf{d} = F\left(\bm{\uptheta}\right) + \textbf{e},
\label{mcmc0}
\end{equation}

where $\textbf{d} = \left(d_1, \ldots, d_N \right) \in \mathbb{R}^N, N \geq 1$ is the measurement data, $F\left(\bm{\uptheta}\right)$ is a deterministic forward model with parameters $\bm{\uptheta}$ and the noise term $\textbf{e}$ lumps all sources of errors. 

In the Bayesian paradigm, parameters in $\bm{\uptheta}$ are viewed as random variables with a posterior pdf, $p\left(\bm{\uptheta} | \textbf{d} \right)$, given by

\begin{equation}
p\left(\bm{\uptheta} | \textbf{d}  \right) = \frac{p \left(\bm{\uptheta}\right) p \left(\textbf{d} | \bm{\uptheta}\right)}{p \left( \textbf{d} \right)} \propto p\left(\bm{\uptheta}\right) L\left(\bm{\uptheta} | \textbf{d}\right),
\label{mcmc1}
\end{equation}

where $L \left(\bm{\uptheta} | \textbf{d}\right) \equiv p \left(\textbf{d} | \bm{\uptheta}\right)$ signifies the likelihood function of $\bm{\uptheta}$. The normalization factor $p \left( \textbf{d} \right) = \int  p\left(\bm{\uptheta}\right) p\left(\textbf{d} | \bm{\uptheta}\right) d\bm{\uptheta}$ is not required for parameter inference when the parameter dimensionality is fixed. In the remainder of this paper, we will thus focus on the unnormalized density $p\left(\bm{\uptheta} | \textbf{d} \right) \propto p \left(\bm{\uptheta}\right) L\left(\bm{\uptheta} | \textbf{d}\right)$.

To avoid numerical over- or underflow, it is convenient to work with the logarithm of $L \left(\bm{\uptheta} | \textbf{d}\right)$ (log-likelihood): $\ell\left(\bm{\uptheta} | \textbf{d}\right)$. If we assume $\textbf{e}$ to be normally distributed, uncorrelated and with known constant variance, $\sigma_e^2$, $\ell\left(\bm{\uptheta} | \textbf{d}\right)$ can be written as

\begin{equation}
\ell\left(\bm{\uptheta} | \textbf{d}\right) = -\frac{N}{2}\log\left(2 \pi\right) - N\log\left(\sigma_e\right) -\frac{1}{2}\sigma_e^{-2}  \sum_{i = 1}^{N} \left[d_i - F_i\left(\bm{\uptheta}\right) \right]^2,
\label{mcmc2}
\end{equation}

where the $F_i\left(\bm{\uptheta}\right)$ are the simulated responses that are compared the $i = 1, \cdots, N$ measurement data, $d_i$.

An exact analytical solution of $p \left(\bm{\uptheta} | \textbf{d}\right)$ is not available for the type of non-linear inverse problems considered herein. We therefore resort to MCMC simulation \citep[see, e.g.,][]{Robert-Casella2004}. More specifically, the DREAM$_{\rm \left(ZS\right)}$ algorithm is used to approximate the posterior distribution. A detailed description of this sampling scheme including a proof of ergodicity and detailed balance can be found in \citet{Vrugt2009} and \citet{Laloy-Vrugt2012}. Multiple contributions in hydrology and geophysics (amongst others) have demonstrated the ability of DREAM$_{\rm \left(ZS\right)}$ to sample target distributions with 50-250 dimensions \citep{Laloy2012,Laloy2013,Linde-Vrugt2013,Laloy2015,Lochbuhler2015}.

\section{Geostatistical simulation results}
\label{results_dr}

Before considering our main goal, that is, solving inverse problems (section \ref{results_inv}), we investigate the performance of our DR approach for  unconditional and conditional geostatistical simulation. This step is important as we aim at producing inversion models that not only honor the data used, but also the spatial statistics of our selected TI. One thus need to make sure that model realizations produced by randomly sampling the constructed low-dimensional base are consistent with the TI.

\subsection{2D channelized aquifer}
\label{channel}

Our first example considers unconditional model generation of a $100 \times 100$ 2D binary and channelized aquifer. The underlying TI is closely related to the classical $250 \times 250$ binary training image introduced by \citet{Strebelle2002} that is routinely used to test new MPS algorithms. Our used TI has some slight additional variability in that the channel rotation angles are allowed to deviate (at most) from -15$\degree$ to +15$\degree$ compared to the rotation angles found in the original TI. The training set used to train the VAECNN contains 80,000 $100 \times 100$ geostatistical realizations generated with the direct sampling (DS) approach as implemented in the DeeSse MPS algorithm \citep{Mariethoz2010b}. Obtaining these 80,000 realizations took 14 hours using eight CPUs in parallel. The choice of 80,000 realizations was dictated by available computational resources. Limited testing further indicated that using 40,000 model realizations for training would very likely not change the results presented below. Training the VAECNN for this example incurred a computational time of approximately 13 hours. As $\textbf{z}$ is 50-dimensional, the achieved compression ratio is 200.

\begin{figure}[H]
	\noindent\hspace{-3.0cm}\includegraphics[width=50pc]{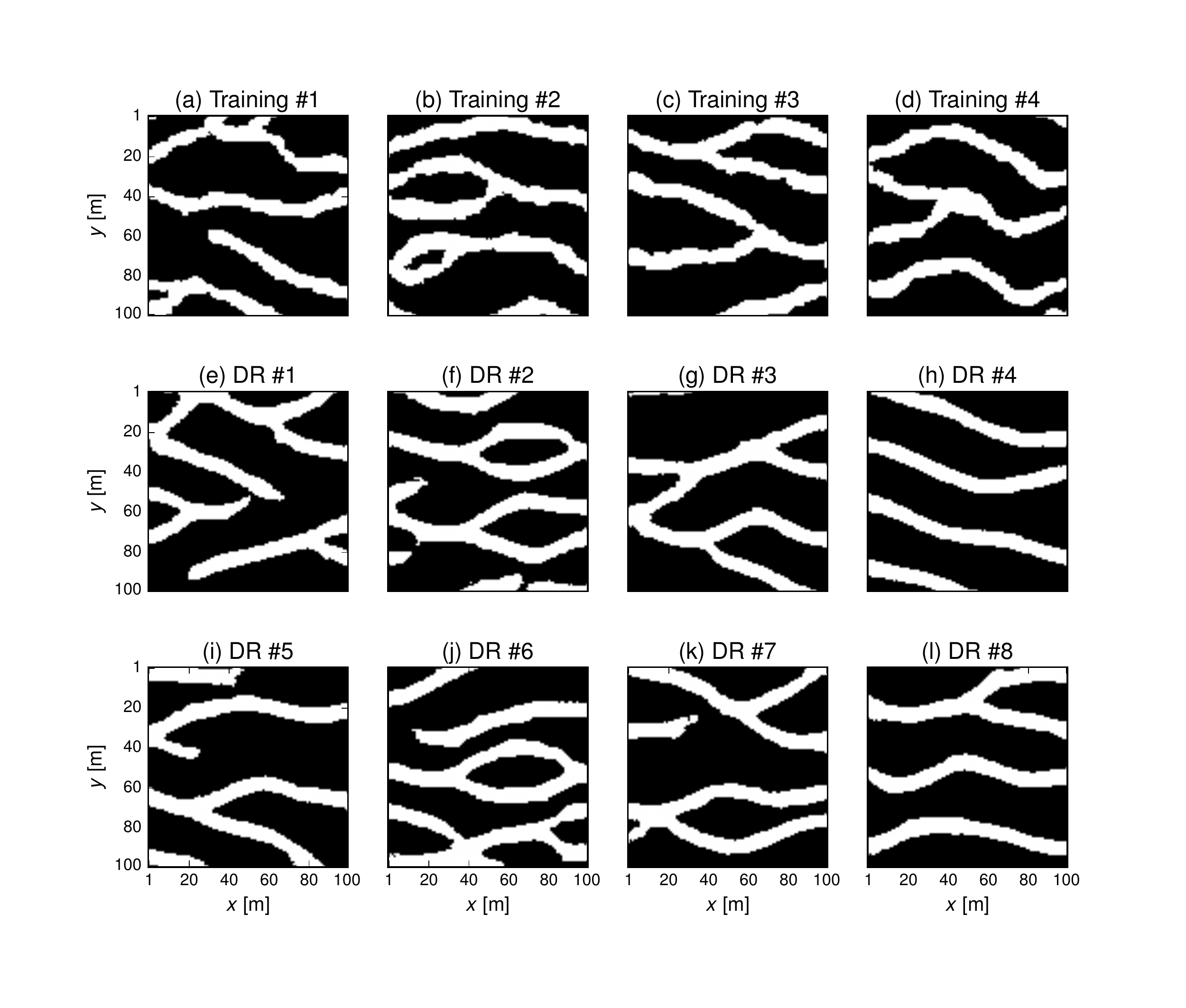}
	\caption{Unconditional model realizations for the 2D binary channelized aquifer TI: (a - d) four (randomly chosen) model realizations from the DS-based training set and (e-l) eight (randomly chosen) model realizations generated with our DR approach that is based on a deep neural network.}
	\label{fig5}
\end{figure}

\begin{figure}[H]
	\noindent\hspace{1cm}\vspace{-1.5cm}\includegraphics[trim=3cm 0 0 5cm,width=35pc]{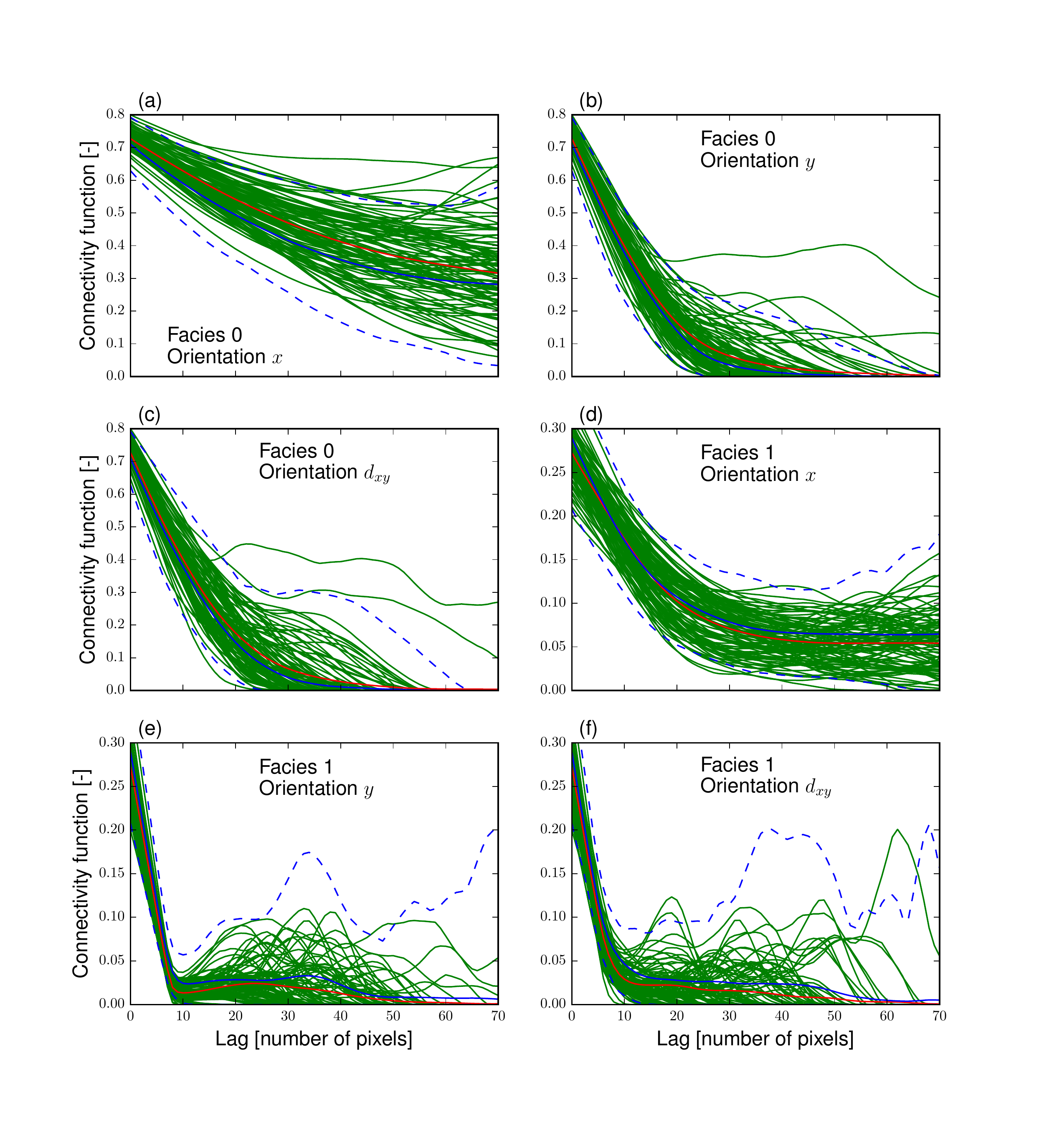}
	\caption{Cluster or connectivity function (CF) for the 2D case study involving a channelized aquifer without direct conditioning data (Figure \ref{fig5}). The blue lines denote the values associated with the training set. The solid blue line indicates the mean while the 2 dashed lines represent the minimum and maximum values at each lag. The green solid lines represent the 100 realizations generated by our DR approach. The red solid line is the mean of these DR-based realizations. The CF is calculated for each facies along directions. The $x$ and $y$ symbols signify the $x$ and $y$-axes, and $d_{xy}$ represents the diagonal direction formed by the 45\degree angle from the $x$-axis.}
	\label{fig6}
\end{figure}

Figure \ref{fig5} presents four (randomly chosen) MPS realizations from the training set, together with eight (randomly chosen) model realizations obtained by our DR approach. The DR-based realizations resemble the training model realizations well, although some artifacts remain in terms of a moderately larger occurrence of broken channels and a slight oversmoothing of the channels. Figure \ref{fig6} displays the corresponding CF metrics. It is observed that the DR-based realizations have similar CF curves as the training set. In terms of space of uncertainty, the ratio of the $\overline{d_{JS}}$ (equation (\ref{spunc1})) of the DR-based realizations to the $\overline{d_{JS}}$ of the training images, $\displaystyle \frac{\overline{d_{JS}}_{(DR)}}{\overline{d_{JS}}_{(TR)}}$, is 1.03. Thus, the DR-based realizations show slightly more between-realization variability than the training set.

\begin{figure}[H]
	\noindent\hspace{-3.0cm}\includegraphics[width=50pc]{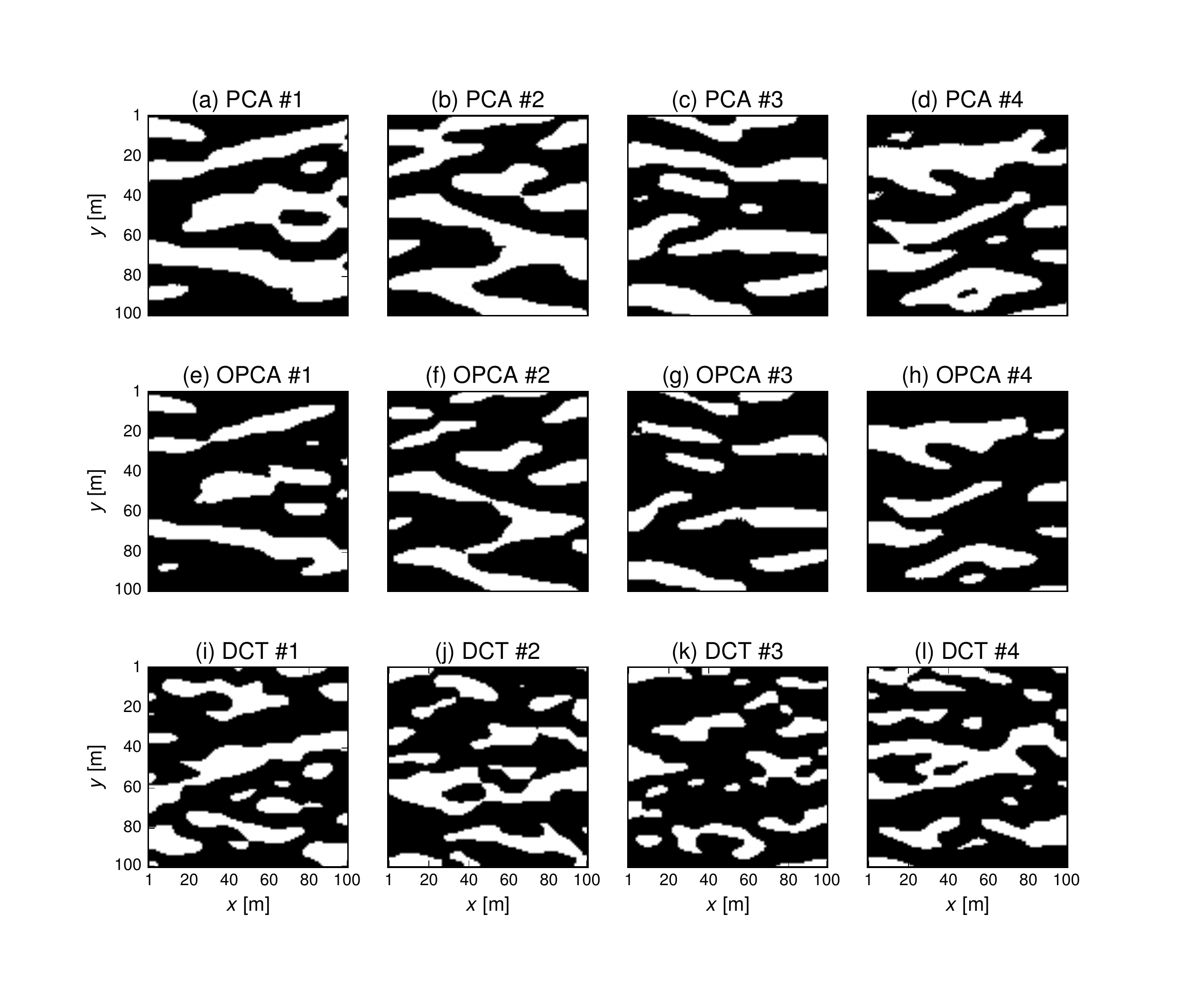}
	\caption{Unconditional model realizations for the 2D binary channelized aquifer TI: (a-d) four (randomly chosen) PCA-based model realizations and (e-h) corresponding OPCA realizations together with (i-l) four (randomly chosen) DCT-based model realizations.}
	\label{fig7}
\end{figure}

Figure \ref{fig7} illustrates corresponding results obtained by the PCA, OPCA and DCT dimensionality reduction methods for the same training set. The PCA and OPCA realizations were derived following \citet{Vo-Durlofsky2014} using 70 random variables. Varying this number in the range 50 - 10000 did not improve generation performance. For DCT, 250 coefficients were used as using a smaller number of coefficients leads to overly degraded models in direct compression mode (compression of existing models). The DCT-based model realizations were then generated by randomly sampling uniform distributions defined by the empirical upper and lower bounds of the 250 maximum (in absolute value) coefficients associated with the training set. The DCT realizations were thresholded such that the facies proportions are close to that of the TI (facies 0: 0.7, facies 1: 0.3). Varying the number of DCT coefficients has no significant impact on generation quality. Clearly, none of these three approaches (PCA, OPCA, DCT) sample from a proper low-dimensional parameter space. In contrast, our proposed approach does a much better job. Note that PCA, OPCA and DCT are far from providing state-of-the-art MPS simulations. Instead, up to now they formed the state-of-the-art in terms of parametric low-dimensional model representations that have been used with advanced inversion methods. The key of our method a low-dimensional parameterization that is able to provide high-quality MPS realizations (compare Figures \ref{fig5} and \ref{fig7}).

\begin{figure}[H]
	\noindent\hspace{-3.0cm}\includegraphics[width=50pc]{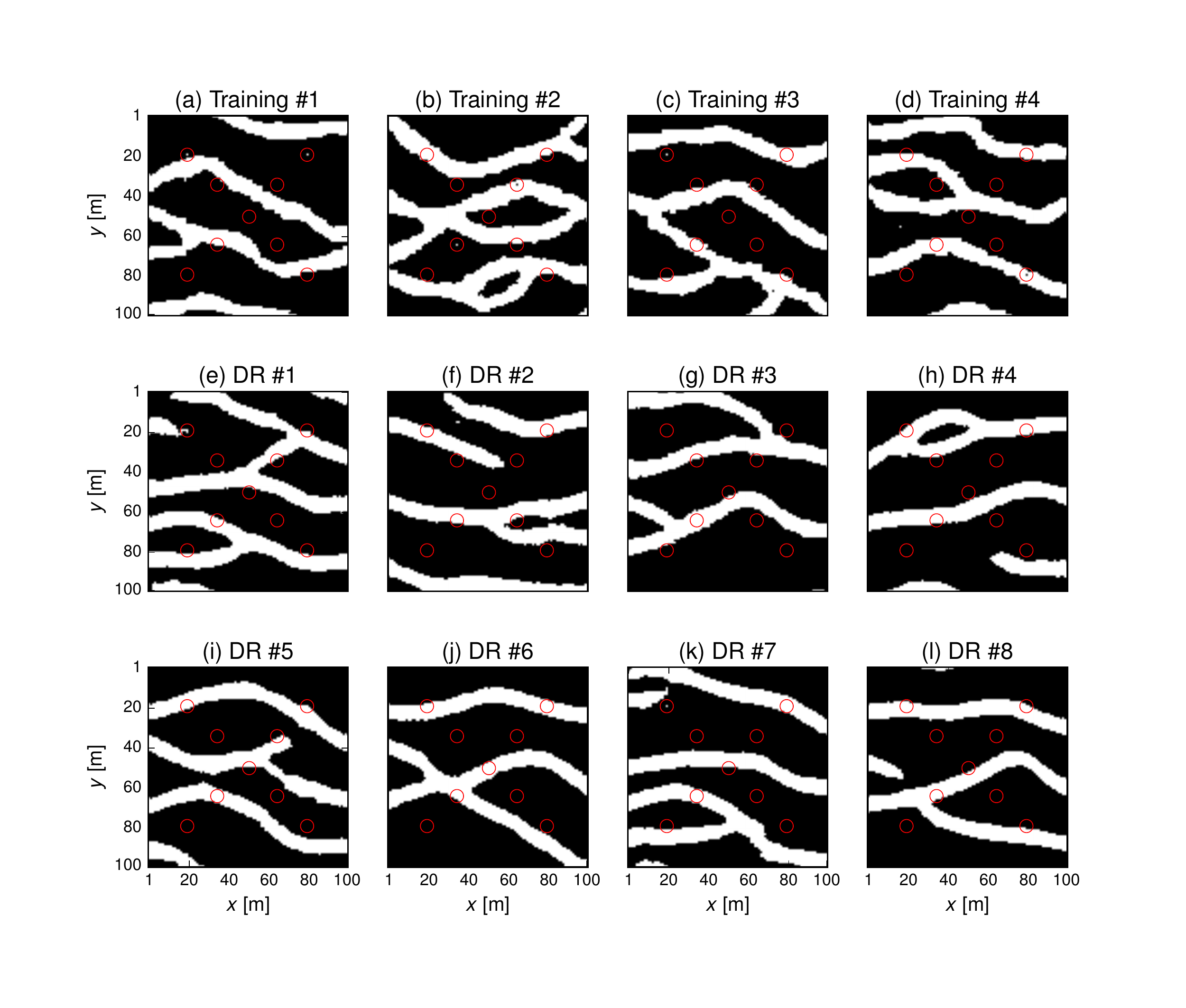}
	\caption{Conditional model realizations for the 2D binary channelized aquifer TI: (a - d) four (randomly chosen) conditioned model realizations from the DS-based training set and (e-l) eight (randomly chosen) conditioned model realizations generated with our DR approach that is based on a deep neural network.}
	\label{fig8}
\end{figure}

Our second example considers a case where direct conditioning data are available. This example is similar to example 1 (Figure \ref{fig5}) except that facies at 9 locations are known (red circles in Figure \ref{fig8}). To handle direct conditioning, our approach requires that the MPS-generated realizations forming the training set honor the conditioning points. Building a training set of 40,000 conditioned models with DS took 7-8 hours using again eight CPUs in parallel. The training of our VAECNN then took 6.5 hours.

\begin{figure}[H]
	\noindent\hspace{1cm}\vspace{-1.5cm}\includegraphics[trim=3cm 0 0 5cm,width=35pc]{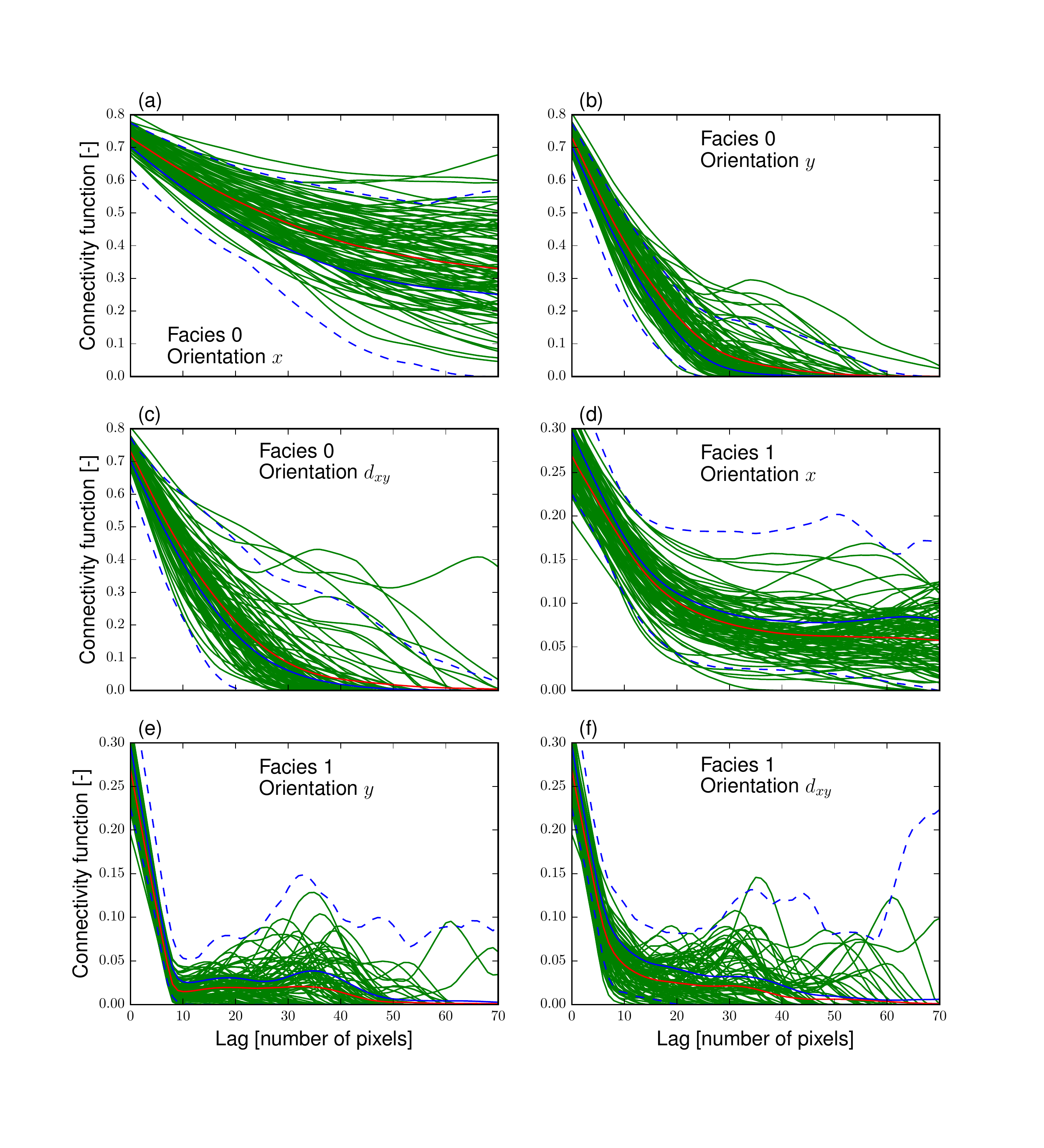}
	\caption{Cluster or connectivity function (CF) for the 2D case study involving a channelized aquifer and 9 direct conditioning data points (Figure \ref{fig8}). The blue lines denote the values associated with the training set. The solid blue line indicates the mean while the 2 dashed lines represent the minimum and maximum values at each lag. The green solid lines represent the 100 realizations generated by our DR approach. The red solid line is the mean of these DR-based realizations. The CF is calculated for each facies along directions. The $x$ and $y$ symbols signify the $x$ and $y$-axes, and $d_{xy}$ represents the diagonal direction formed by the 45\degree angle from the $x$-axis.}
	\label{fig9}
\end{figure}

A set of (randomly chosen) training and DR-generated model realizations are shown in Figure \ref{fig8}. A similar generation performance is visually observed as for example 1 (Figure \ref{fig5}) and the DR-based realizations show again similar CF curves as the training images (Figure \ref{fig9}). However, the achieved space of uncertainty is 12\% smaller than that of the conditioned training set: $\displaystyle \frac{\overline{d_{JS}}_{(DR)}}{\overline{d_{JS}}_{(TR)}} = 0.88$.

The conditioning accuracy is assessed by analyzing 1000 random model realizations. We find that 68\% of the model realizations honor all of the 9 prescribed facies and that 97\% of the model realizations contain at most one mismatching conditioning datum. Also, the more common facies 0 is more frequently honored (98\%) than the less frequent facies 1 (91\%). This is less than for our DS-generated training set that always honor the nine conditioning points. Nevertheless, the training realizations (Figures \ref{fig8}a-d) often contain 1 or 2 isolated conditioning point(s) within an homogeneous zone of the other facies. This type of artifact is less frequent in the DR-based model realizations (Figures \ref{fig8}e-l). The main reason for the conditioning errors is the sequential relooping used for producing a realization (see section \ref{methods_architecture_details}). Without relooping, the model realizations present more frequently broken channels and/or small groups of isolated pixels (not shown), but the percentage of realizations that fully honor the 9 conditioning data is 92\% (instead of 68\% with relooping) and 100\% (instead of 97\% with relooping) of the realizations has at most a single erroneous datum.

\subsection{3D model}
\label{3d_gen}

We now turn our attention to the generation of 3D model realizations. The selected TI is the Maules Creek valley alluvial aquifer available from \url{http://www.trainingimages.org/training-images-library.html}. For the unconditional case, a training set of 19,500 $30 \times 32 \times 27$ model realizations was built using the recent graph cuts (GC) patch-based MPS algorithm by \citet{Li2016}. Running 8 different GC instances simultaneously, this took approximately 6 hours. The GC method was selected because it is much faster than DS. In addition, training our VAECNN took about 7 hours.

The convolutional layers in our VAECNN are designed for 2D images only, but with possibly different color channels. To generate 3D images, one can use as many channels as there are horizontal layers in the model (27). This does not explicitly account for patterns in the vertical direction since the convolutional operations are performed in the horizontal plane only. Nevertheless, equation (\ref{dnn8}) penalizes reconstructed models that deviate from the training set in all three spatial directions. 

\begin{figure}[H]
	\noindent\hspace{-2.0cm}\includegraphics[width=45pc]{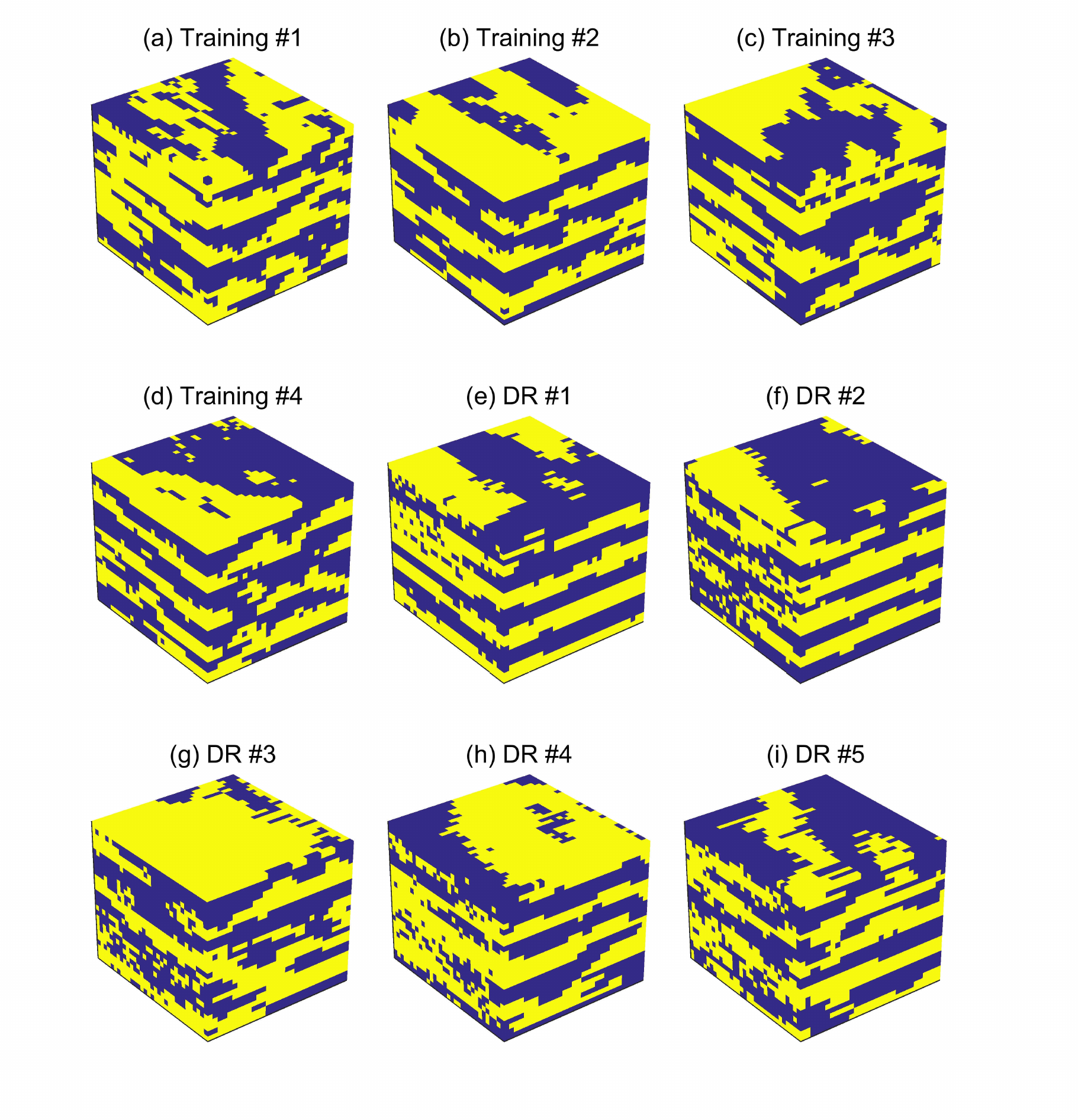}
	\caption{Unconditional model realizations for the 3D Maules Creek binary aquifer TI: (a - d) four (randomly chosen) model realizations from the MPS-based training set and (e-l) five (randomly chosen) model realizations generated with our DR approach that is based on a deep neural network.}
	\label{fig10}
\end{figure}

\begin{figure}[H]
	\noindent\hspace{0.5cm}\vspace{-1.5cm}\includegraphics[trim=5cm 0 0 5cm,width=40pc]{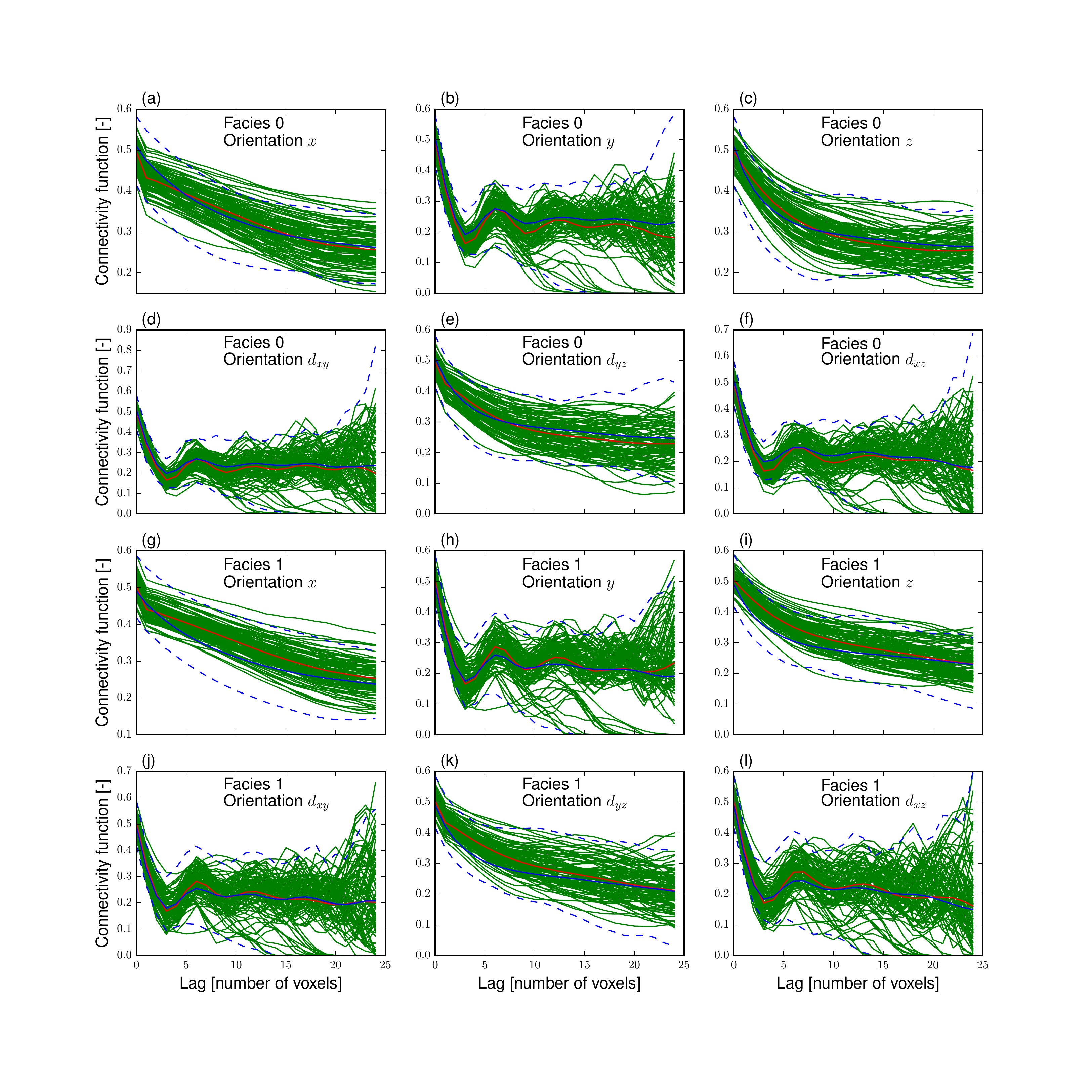}
	\caption{Cluster or connectivity function (CF) for the 3D case study involving the Maules Creek binary aquifer without direct conditioning data (Figure \ref{fig10}). The blue lines denote the values associated with the training set. The solid blue line indicates the mean while the 2 dashed lines represent the minimum and maximum values at each lag. The green solid lines represent the 100 realizations generated by our DR approach. The red solid line is the mean of these DR-based realizations. The CF is calculated for each facies along directions. The $x$, $y$ and $z$ symbols signify the $x$, $y$ and $z$-axes. The $d_{xy}$, $d_{yz}$ and $d_{xz}$ symbols represent the diagonal direction formed by the 45\degree angle from the $x$-axis in the $xy$ plane, from the $y$-axis in the $yz$ plane and from the $x$-axis in the $xz$ plane, respectively.}
	\label{fig11}
\end{figure}

Figure \ref{fig10} compares (randomly chosen) training and DR-based realizations. As the model domain contains $30 \times 32 \times 27 = 25,920$ voxels, the compression ratio is $25920/50 \approx 518$. The proposed DR approach creates realizations that are similar to those in the training set, although isolated voxels of each facies are over-represented. The realizations also reproduce well the CF metrics of the training set (Figure \ref{fig11}).

\begin{figure}[H]
	\noindent\hspace{-3.5cm}\includegraphics[width=50pc]{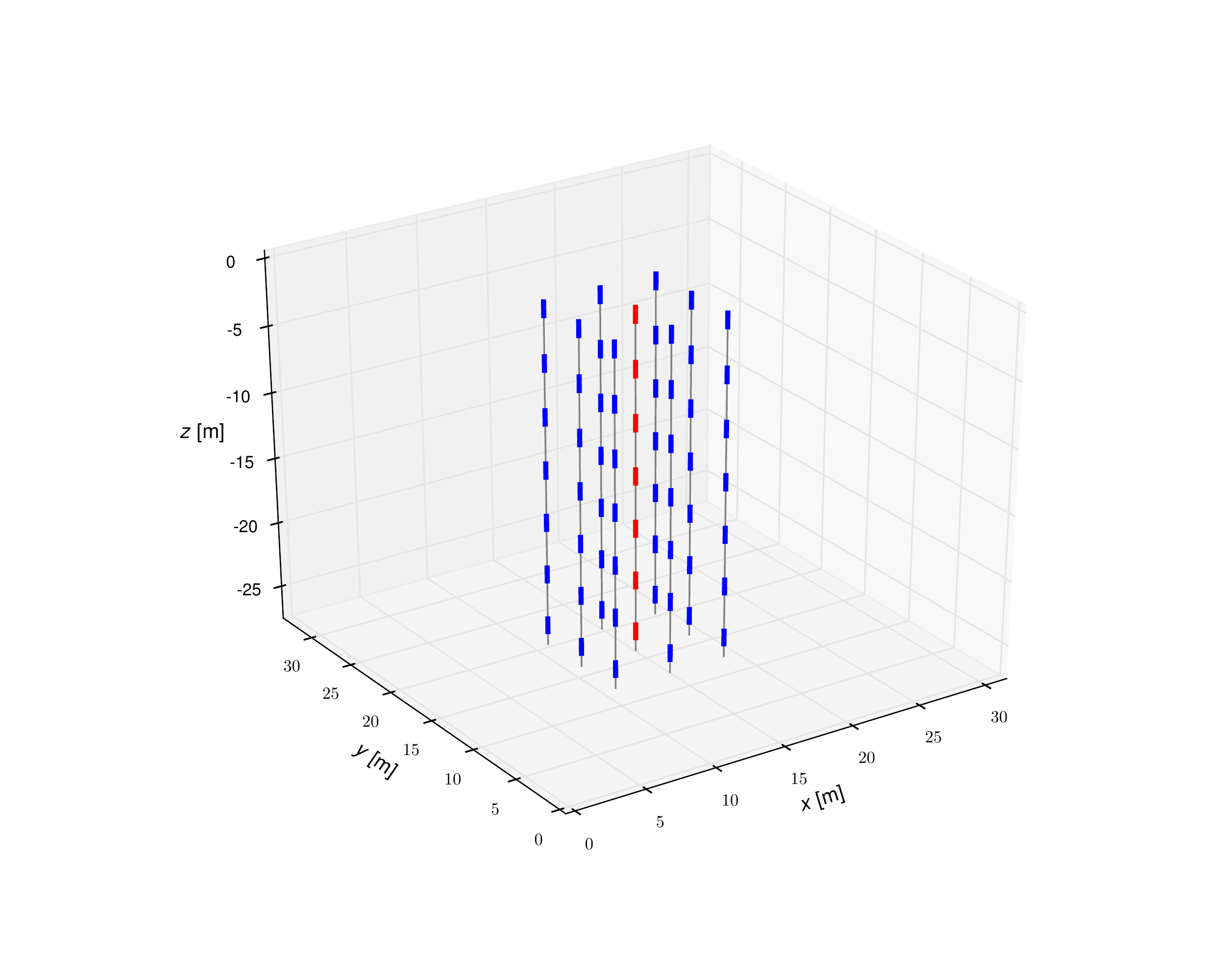}
	\caption{Pumping and observation locations in the synthetic 3D hydraulic tomography setup. The vertical gray lines represent the multilevel pumping and observation wells. The red line segments in the central well are the pumping locations and the blue line segments in the other wells are the head measurement locations.}
	\label{fig12}
\end{figure}

We also considered the same 3D example with 56 direct conditioning points. These points correspond to the locations of the screens along the vertical multi-level piezometers used in our synthetic 3D inverse problem described in section \ref{inv3d} (Blue line segments in Figure \ref{fig12}). A GC-based training set containing 19,000 conditioned model realizations was used for training our VAECNN. Running again 8 different GC instances simultaneously, building the training set took approximately 14 hours. Similarly as for the unconditional case, training then lasted for 7 hours. Using default algorithmic settings, these original GC simulations jointly honor all of the 56 conditioning data only 6\% of the times but contain at most 10 mismatching point 94\% of the times. The DR-based realizations are again visually close to the training set (not shown), with a much smaller tendency to overproduce isolated voxels of each facies than for the unconditional case (not shown) and CF statistics that closely match those of the training set (not shown). Furthermore, our DR approach is found to condition at a slightly higher level than the selected GC algorithm. Among 1000 random DR-based realizations, 7\% of the realizations honor all of the 56 prescribed facies while 98\% of the realizations contain a most 10 mismatching points. We attribute these slightly superior statistics compared to the training set to a combination of randomness and complex bias in the dimensionality reduction. Overall, our VAECNN is found to condition equally well as the GC algorithm used to produce the training set.

\section{Inverse problems}
\label{results_inv}
\subsection{Case study 1: 2D steady-state flow}
\label{inv2d}

Our first inversion case study considers steady-state flow within a channelized aquifer. The 100 $\times$ 100 aquifer domain lies in the $x-y$ plane with a grid cell size of 1 m and a thickness of 1 m. Channel material and matrix material (see Figures \ref{fig14}a and \ref{fig15}a) are assigned hydraulic conductivity values of 1 $\times$ 10$^{-2}$ m/s and 1 $\times$ 10$^{-4}$ m/s, respectively. Steady state groundwater flow is simulated using MODFLOW 2005 \citep{Harbaugh2005} assuming no flow boundaries at the upper and lower sides and fixed head boundaries on the left and right sides of the domain to ensure a lateral head gradient of 0.01 (-), with water flowing in the $x$-direction. A pumping well extracting 0.001 m$^3$/s is located at the center of the domain. Simulated heads are collected at 49 locations that are regularly spread over the domain (Figures \ref{fig14}a and \ref{fig15}a). These data were then corrupted with a Gaussian white noise using a standard deviation of 0.02 m. For the selected white noise realization, the measurement data have a root-mean-square-error (RMSE) of 0.0200 m. The corresponding SNR, defined as the ratio of the average RMSE when drawing prior realizations with our DR algorithm to the noise level is 25. No direct conditioning (facies) data are used.

We first verify whether a model realization belonging to the 50-dimensional manifold can be retrieved by the inversion process. To do so, the reference model (Figure \ref{fig14}a) was randomly generated using our DR approach. The DREAM$_{\rm \left(ZS\right)}$ was ran in parallel, using 8 interacting Markov chains distributed over 8 CPUs. Uniform priors in $\left[-5,5\right]$ were selected for the 50 DR variables, $\bm{\uptheta}$. This might seem to contradict the theory behind our VAECNN which is based on calibration to a standard normal distribution for $p\left(\textbf{z}\right)$. However, using a standard normal prior for $\bm{\uptheta}$ was found to restrict the sampled model space too much for the inversion process to produce model realizations that in average fit the data to prescribed noise level. Our explanation for this phenomena is that with 50 DR variables the MCMC simulation needs sufficient freedom in the explored model space to eventually produce consistent models. Also, and equally important, is the observation that after training using equations (\ref{dnn8}-\ref{dnn9}), the quality of the generated realizations by our VAECNN is similar when using a zero-mean uniform or a standard normal distribution for $p\left(\textbf{z}\right)$. From a pragmatic point of view, the zero-mean uniform prior can thus be used.

\begin{figure}[H]
	\noindent\hspace{-2.0cm}\includegraphics[width=45pc]{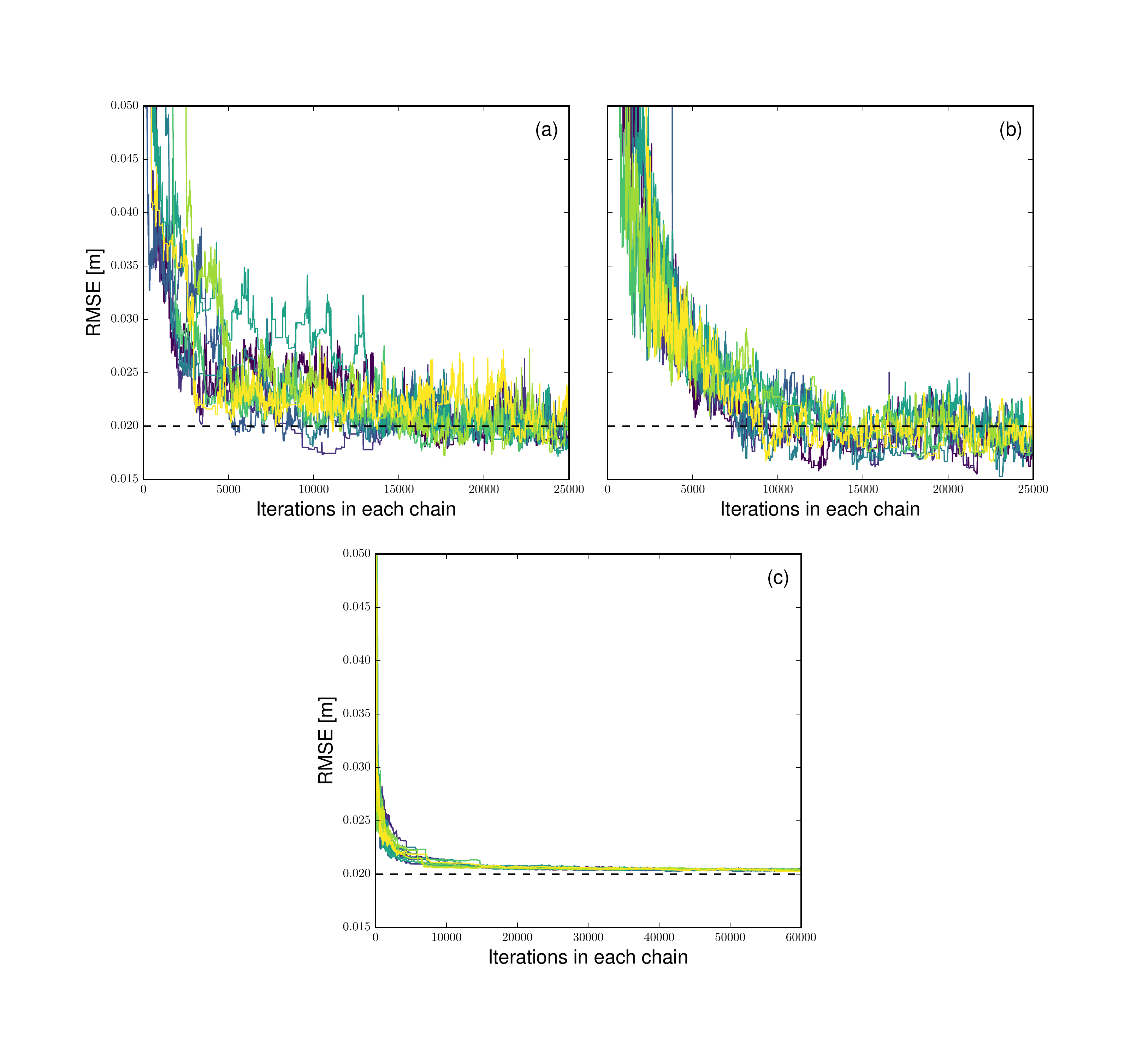}
	\caption{Trace plot of the sampled RMSE values by (a) the 8 Markov chains evolved by DREAM$_{\rm \left(ZS\right)}$ (colored lines) for the first test (DR-generated true model) of inverse case study 1, (b) the 8 Markov chains evolved by DREAM$_{\rm \left(ZS\right)}$ (colored lines) for the second test (DS-generated true model) of inverse case study 1 and (c) the 16 Markov chains evolved by DREAM$_{\rm \left(ZS\right)}$ (colored lines) for the inverse case study 2. The dashed black line denotes the true RMSE value.}
	\label{fig13}
\end{figure}

\begin{figure}[H]
	\noindent\hspace{-2.0cm}\includegraphics[width=42pc]{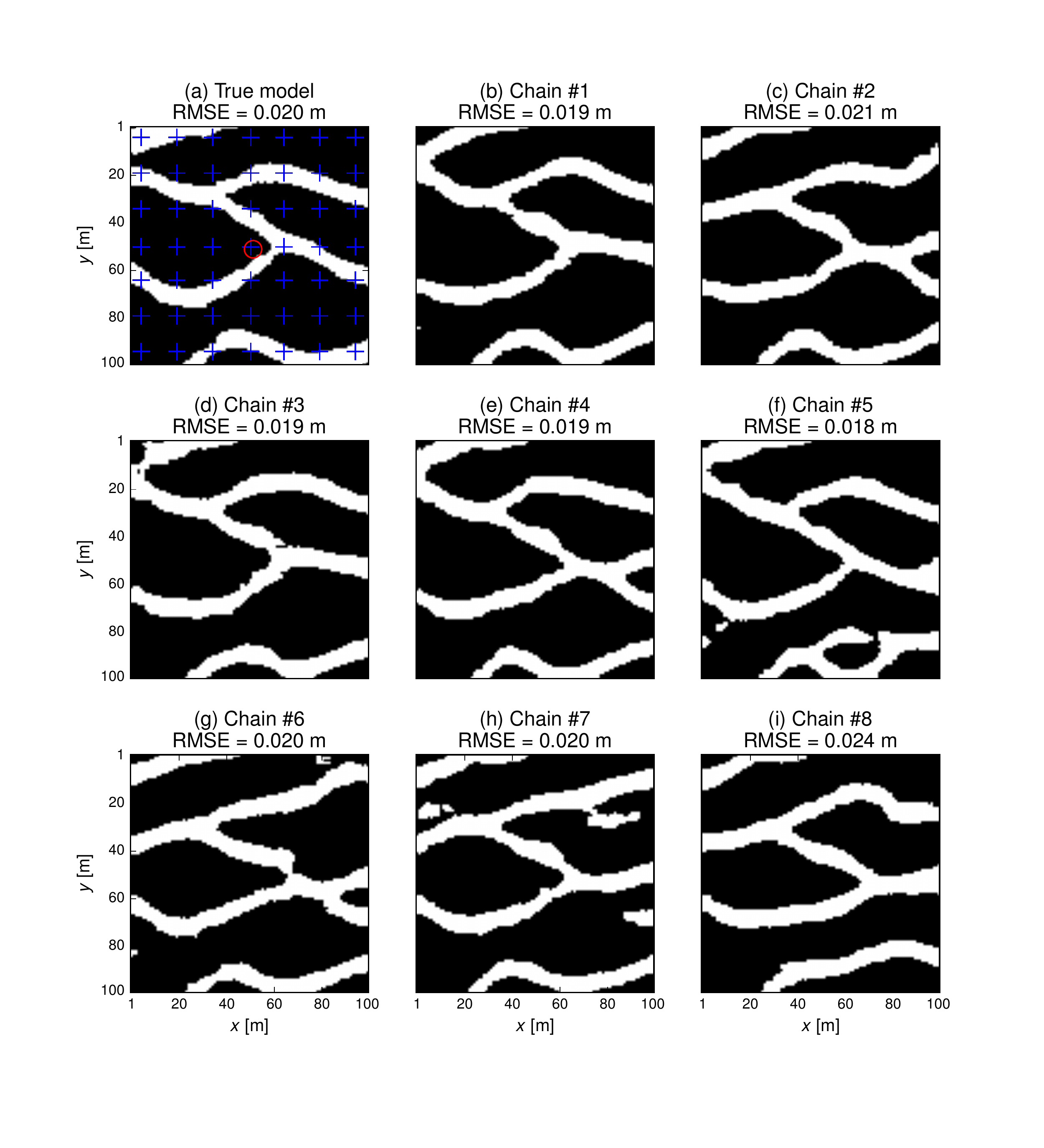}
	\caption{(a) True model and (b-i) states of the 8 Markov chain evolved by DREAM$_{\rm \left(ZS\right)}$ after 25,000 iterations per chain for the first test (DR-generated true model) of inverse case study 1. The red circle and blue crosses in subfigure (a) mark the location of the pumping well and the piezometers, respectively.}
	\label{fig14}
\end{figure}

\begin{figure}[H]
	\noindent\hspace{-2.0cm}\includegraphics[width=42pc]{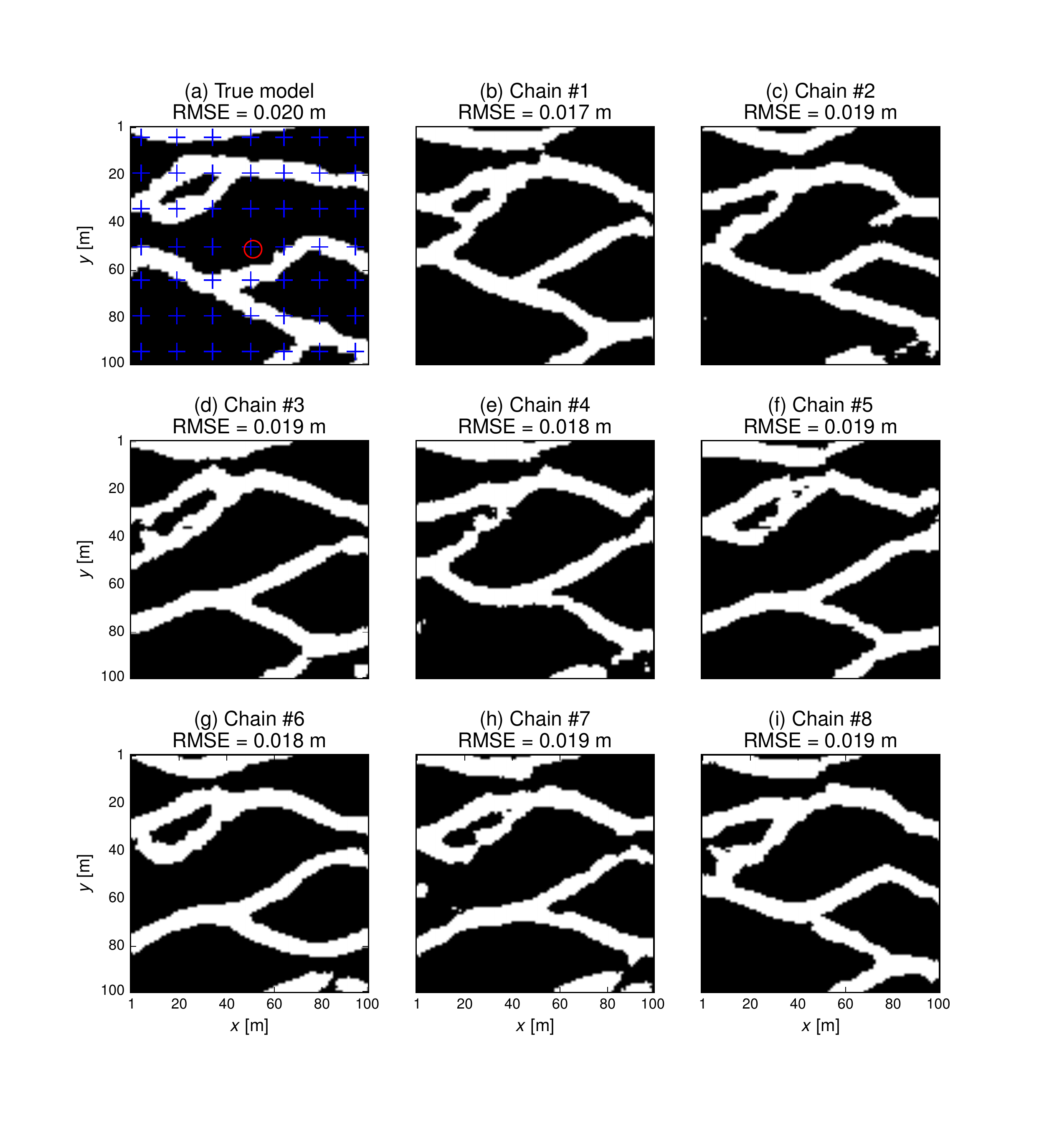}
	\caption{(a) True model and (b-i) states of the 8 Markov chains evolved by DREAM$_{\rm \left(ZS\right)}$ after 25,000 iterations per chain for the second test (MPS-generated true model) of inverse case study 1. The red circle and blue crosses in subfigure (a) mark the location of the pumping well and the piezometers, respectively.}
	\label{fig15}
\end{figure}

The chains start to jointly sample the posterior distribution, $p\left(\bm{\uptheta} | \textbf{d}  \right)$,  after a (serial) total of 120,000 iterations, that is, 15,000 parallel iterations  per chain (Figure \ref{fig13}a). The sampled realizations closely resemble the true model and the posterior variability is overall small (Figure \ref{fig14}). To assess sampling accuracy, we calculate over the last 10,000 posterior realizations in each chain the average fraction of pixels with facies identical to that of the true model, $f_{\rm PO}$, which is 0.86. The same quantity for the prior, $f_{\rm PR}$, can be directly calculated as $0.7 \times 0.75 + 0.3 \times 0.25 = 0.60$ with 0.7 and 0.3 the prior fractions of each facies in the TI and 0.75 and 0.25 the fractions of each facies in the true model. The $\displaystyle\frac{f_{\rm PO}}{f_{\rm PR}}$ ratio will be used to compare the examples considered in this section. Here $\displaystyle\frac{f_{\rm PO}}{f_{\rm PR}} = \frac{0.86}{0.60} = 1.43$. After a total of 200,000 MCMC iterations, the \citet{Gelman-Rubin1992} convergence diagnostic, $\hat{R}$, is satisfied (i.e., $\hat{R} \leq 1.2$) for 37 out of the 50 sampled parameters \citep[see, e.g.,][for details about the use of $\hat{R}$ with DREAM$_{\rm \left(ZS\right)}$]{Laloy2015}. The MCMC sampling should therefore be continued for a longer time for (official) full exploration of the posterior distribution.

Our second and more comprehensive test uses a true model that was generated by the DS algorithm. If the DR-representation is inappropriate, this true model might not be part of the 50-dimensional DR space. Identical inversion settings were used as in the previous example. Here the 8 chains start to jointly sample $p\left(\bm{\uptheta} | \textbf{d}  \right)$ after some 12,500 iterations per chain (Figure \ref{fig13}b). The posterior model realizations look visually close to the reference model, though posterior variability appears to be a bit larger than for the previous case (Figure \ref{fig15}). The associated $f_{\rm PO}$ and $f_{\rm PR}$ values are 0.78 and $0.7 \times 0.71 + 0.3 \times 0.29 = 0.59$, respectively, which gives a $\displaystyle\frac{f_{\rm PO}}{f_{\rm PR}}$ ratio of 1.32. The results displayed in Figure \ref{fig15} inspire confidence that the proposed DR-based inversion approach can retrieve consistent posterior models. With respect to posterior exploration, formal convergence is yet to be declared after a total of 200,000 serial iterations, with the $\hat{R}$ criterion being satisfied for 29 out of the 50 dimensions of $\bm{\uptheta}$. Full posterior exploration therefore requires longer chains.

\newpage

\begin{figure}[H]
	\noindent\hspace{-2.0cm}\includegraphics[width=42pc]{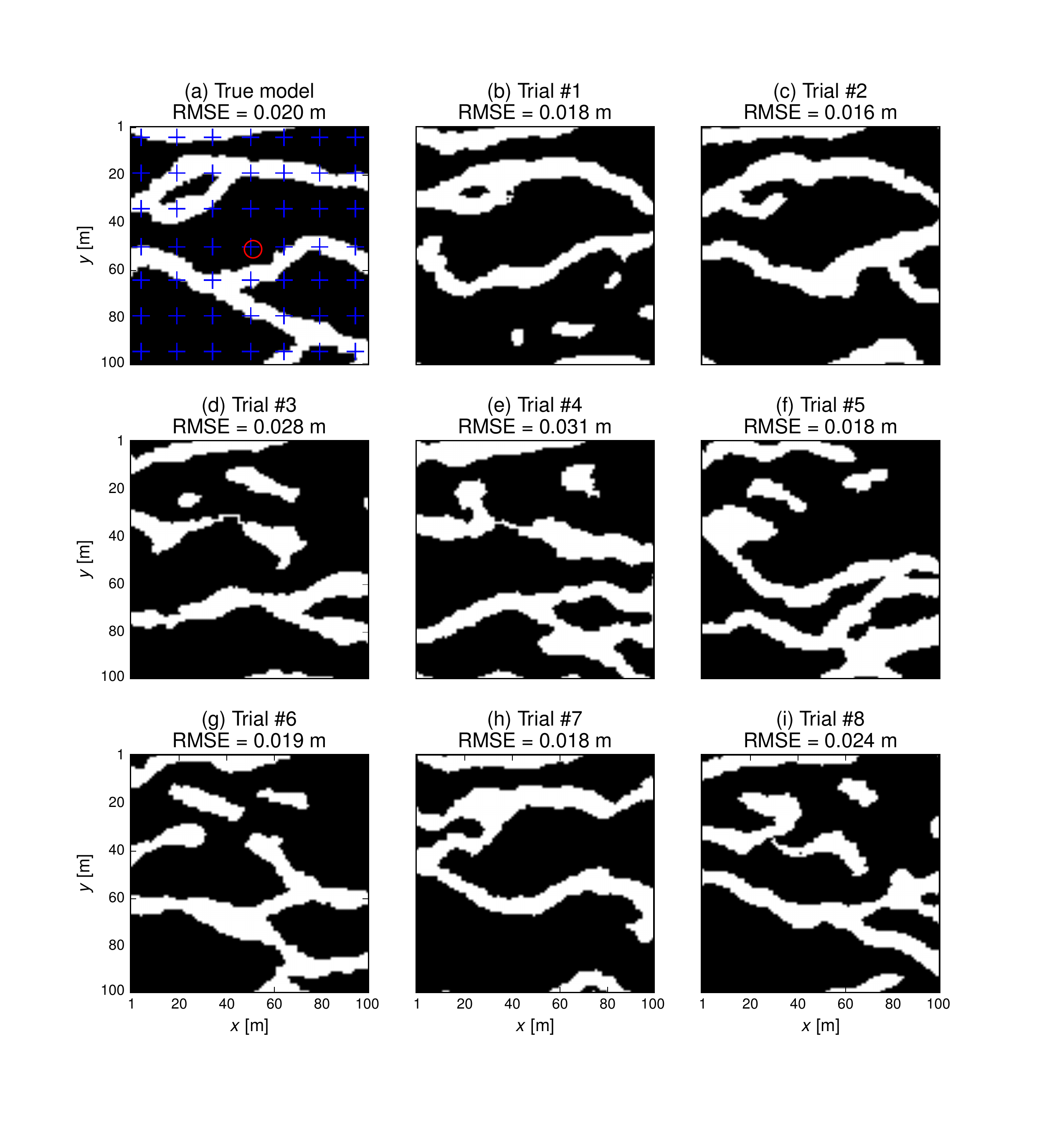}
	\caption{(a) True model and (b-i) states of the 8 independent SGR trials after 25,000 iterations per trial for the second test (MPS-generated true model) of inverse case study 1. The red circle and blue crosses in subfigure (a) mark the location of the pumping well and the piezometers, respectively.}
	\label{fig16}
\end{figure}

To benchmark our approach against state-of-the-art MPS inversion, we performed eight independent SGR runs in parallel for the same case study as above and a similar CPU budget of 25,000 MCMC iterations per run. Moreover, the SGR algorithmic settings are optimal (for the considered type of application) as defined in \citet{Laloy2016}. It is observed that 5 SGR trials out of 8 fit the data to the appropriate noise level after less than 10,000 iterations (not shown). Three trials even reach the targeted 0.02 m misfit after less than 3000 iterations (not shown). Yet three other SGR trials do not sample the posterior distribution after the 25,000 iterations (trials \#3, \#4 and \#8 in Figure \ref{fig16}). Furthermore, all SGR trials end up in a relatively narrow local optimum (not shown) while none of them find a model that is visually similar to the true model (Figure \ref{fig16}). Many of the realizations sampled by SGR are also degraded compared to the prior model (see top row of Figure \ref{fig5}), with isolated patches and broken channels (Figure \ref{fig16}). This translates into a quite lower $\displaystyle\frac{f_{\rm PO}}{f_{\rm PR}}$ ratio than for our DR-based approach: 1.08 (the average $f_{\rm PO}$ over the 8 SGR trials is 0.64). We thus conclude that for the considered example, our DR approach is a superior alternative to SGR.

\subsection{Case study 2: 3D transient hydraulic tomography}
\label{inv3d}

\begin{figure}[H]
	\noindent\hspace{-1.5cm}\includegraphics[width=42pc]{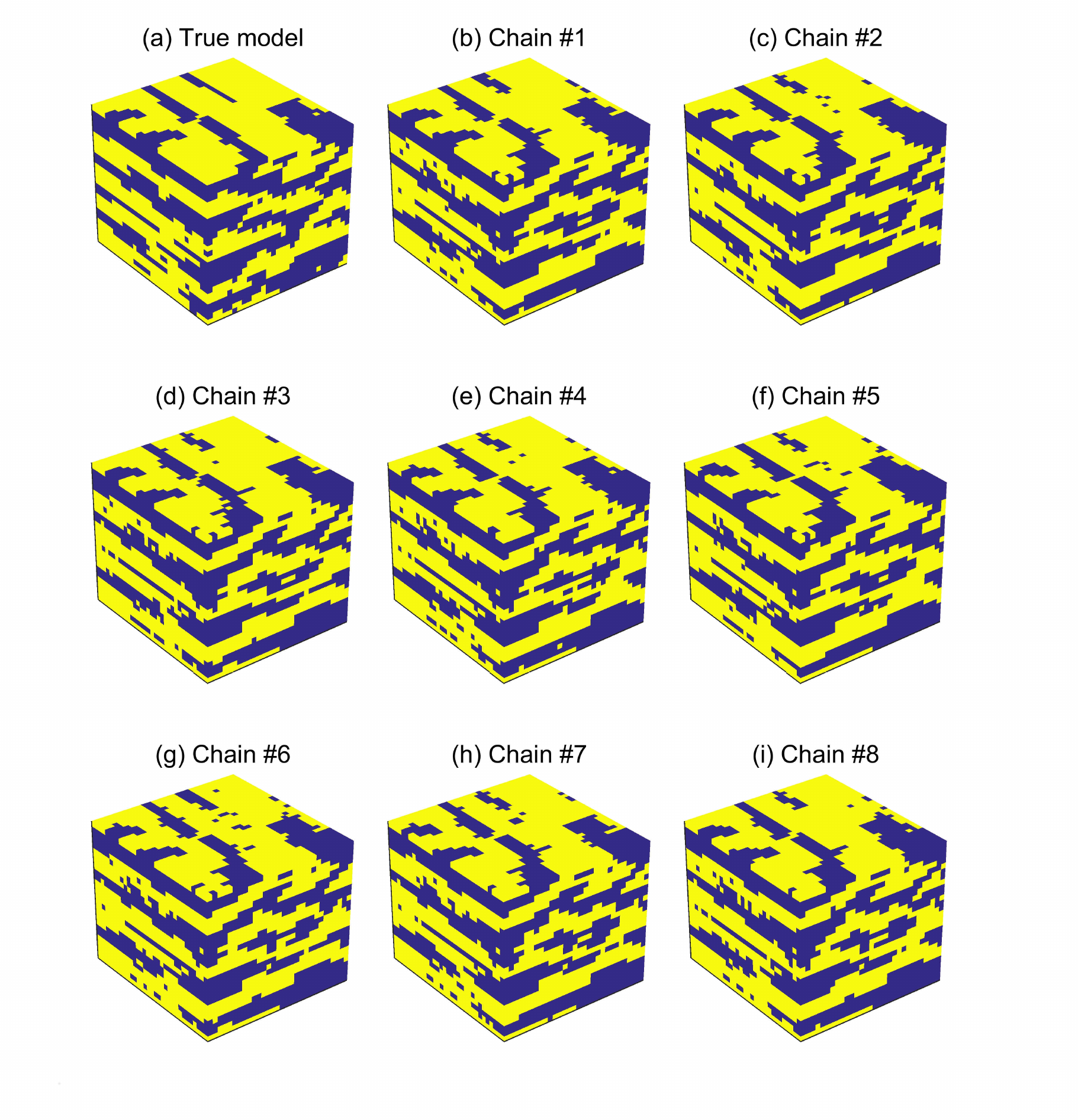}
	\caption{a) True model and (b-i) states of the first 8 (out of 16) Markov chains evolved by DREAM$_{\rm \left(ZS\right)}$ after 60,000 iterations per chain for the inverse case study 2.}
	\label{fig17}
\end{figure}

Our last case study focuses on 3D transient hydraulic tomography \citep[e.g.,][]{Cardiff2013}. Transient head variations caused by discrete multilevel pumping tests were simulated using MODFLOW2005 within a 3D confined aquifer of size 30 $\times$ 32 $\times$ 27 with a voxel size of 1 m $\times$ 1 m $\times$ 1 m. This aquifer consists of a part of the Maules Creek TI discussed in section \ref{3d_gen}. The true model is depicted in Figure \ref{fig17}a. The multilevel discrete pumping setup consists of a 27-m deep, central multi-level well in which water is sequentially extracted every 4 m along a 1-m long screen (at depths of 3, 7, 11, 15, 19, 23 and 27 m respectively) during 30 minutes at a rate of 20 liters/min. The locations of this multilevel pumping well and those of the 8 surrounding multilevel observation wells are displayed in Figure \ref{fig12}. For each pumping sequence, drawdown are recorded every 4 m along a 1-m long screen in the 8 multi-level piezometers (Figure \ref{fig12}). For each drawdown curve, data acquired at the same four measurement times were retained leading to a total of $8 \times 7 \times 7 \times 4 = 1568$ measurement data. These measurement times were considered to be the four most informative ones after visual inspection of several drawdown curves (not shown). These data were corrupted with a Gaussian white noise using again a standard deviation of 0.02 m, which induced for the selected white noise realization a RMSE of 0.0200 m. The associated SNR is 6. In addition, the facies of the 56 locations (voxels) where pressure head is measured (blue line segments in Figure \ref{fig12}) are known and used as direct conditioning data in the inversion.

For this case study, the DREAM$_{\rm \left(ZS\right)}$ sampler evolves 16 Markov chains in parallel using 16 CPUs. Uniform priors in $\left[-5,5\right]$ were again selected for the 50 dimensions of $\bm{\uptheta}$. Within the allowed computational budget of 60,000 iterations per chain, the 16 chains converge towards a data misfit in the range of 0.0202 m - 0.0204 m (Figure \ref{fig13}c). This interval is close to the target level of 0.0200 m. This indicates that the posterior mode has not been sampled yet. The variability among the sampled models is rather small (Figure \ref{fig17}). The latter is likely due to the combination of two factors: (1) the peakedness of the likelihood function (equation (\ref{mcmc2})) caused by the large number of measurement data, and (2) the difficulties encountered by the sampling algorithm for exploring this complex target distribution. Notwithstanding, the sampled models (Figure \ref{fig17}b-i) are visually close to the true model (Figure \ref{fig17}a).

\section{Discussion}
\label{discussion}

\begin{figure}[H]
	\noindent\hspace{0cm}\includegraphics[width=32pc]{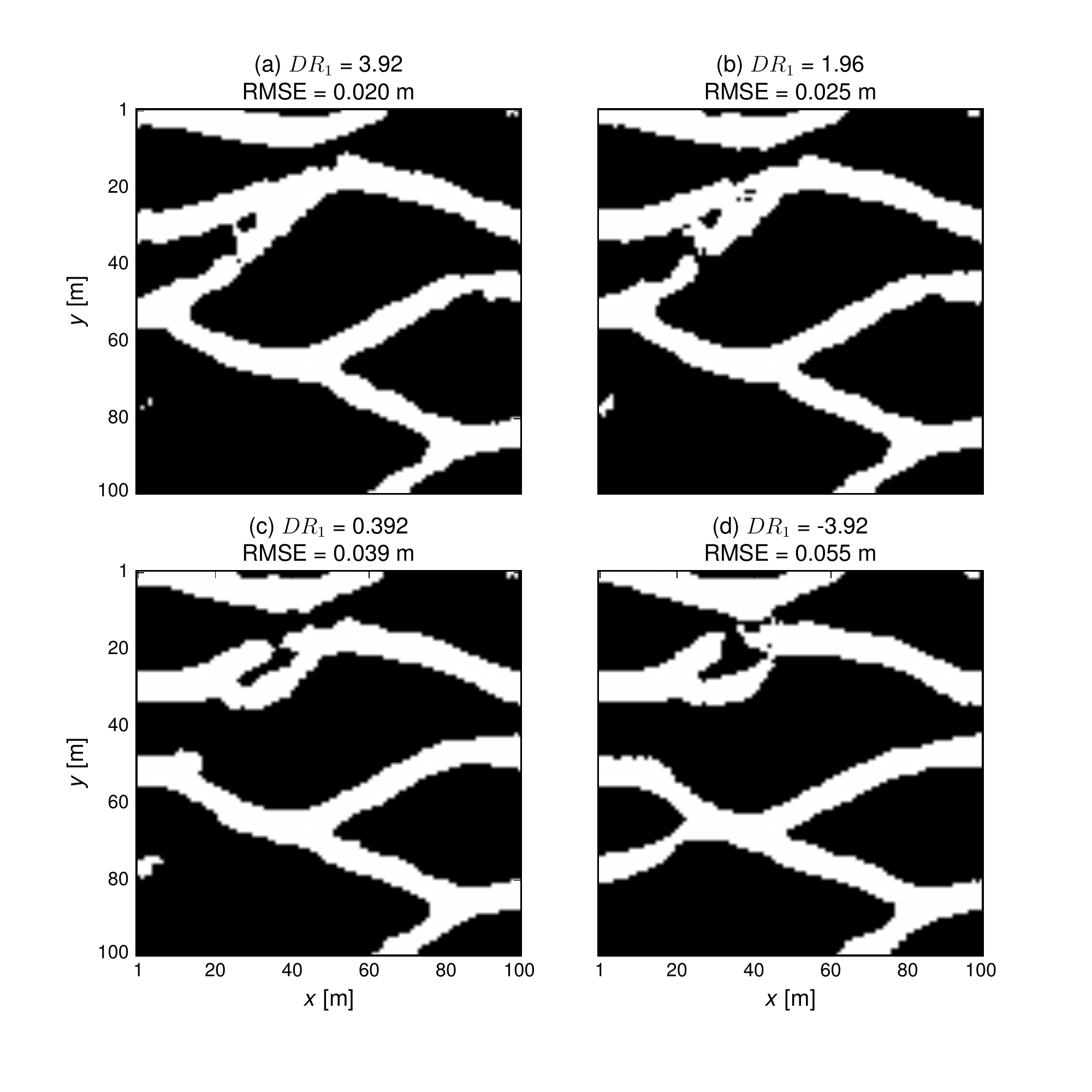}
	\caption{Model realizations and associated RMSE obtained when modifying a single low-dimensional dimension, $DR_1$, of a posterior solution for case study 1: (a) reference (posterior) realization, (b) realization and RMSE corresponding to the multiplication of $DR_1$ by 0.5, (c) realization and RMSE corresponding to the multiplication of $DR_1$ by 0.1 and (d) realization and RMSE corresponding to the multiplication of $DR_1$ by -1.}
	\label{fig18}
\end{figure}

Our proposed DR approach that is based on a deep neural network works well for the considered inversion of 2D steady-state flow data. It also provides useful solutions to our inverse problem involving 3D transient flow. However, further improvements are needed. As for most compression approaches, our dimensionality reduction scheme builds a low-dimensional representation where the value in each dimension influences every pixel/voxel of the generated realization in the original space. This ``non-localization" poses problems for inverse problems involving categorical fields. Indeed, \citet{Laloy2016} have shown for SGR that iterative perturbation of a small well defined fraction of the current model works much better than slightly perturbing the whole model at every step. The same logic applies to compression-based inversion. Indeed, Figure \ref{fig18} demonstrates for our first inversion case study that (1) changing the value of a single low-dimensional parameter influences the whole model realization domain and (2) even a small variation in the model realization can cause a large upward shift in RMSE (e.g., compare Figures \ref{fig18}a and \ref{fig18}c). This implies that our proposed inversion approach can have troubles to converge when applied to rich datasets with low error (e.g., $\leq$ 0.01 m for measured piezometric heads). We expect that a yet to be developed ``localized" DR approach would perform better.

There are other aspects that would benefit from improvements. First, our DR approach is not totally accurate for direct conditioning. Second, several tens of thousands of MPS-generated training models are used to build it. This might not be practical when MPS-generation is computationally intensive. Third, 3D model generation capabilities could be enhanced by considering 3D convolutional layers instead of tricking 2D convolutional layers as done herein. Fourth, using a different reconstruction loss function than equation (\ref{dnn8}) the approach could be extended to categorical TIs with more than two facies and to continuous TIs. Lastly, it may be considered as a disadvantage that our approach produces continuous realizations and thus requires thresholding to create categorical fields.

Because of computational constraints, the impact of using a different number of dimensions for $\textbf{z}$ on geostatistical simulation quality was not studied extensively. Yet limited testing revealed that for the considered 2D channelized TI and associated model domain size using a 25-dimensional $\textbf{z}$ does not produce sufficient variability in the realizations (not shown). In contrast, compared to a 50-dimensional $\textbf{z}$ using 100 dimensions induces fewer broken channels and thus provides slightly better results (not shown). For the considered 3D Maules Creek aquifer TI, using a 25-dimensional $\textbf{z}$ was found to perform equally well as a 50-dimensional $\textbf{z}$ (not shown). These findings warrant further investigations.

We would like to stress that if no hard thresholding is done at the end of our model generation process, then the derivatives of the pixels/voxels of the generated model, $\textbf{X}$, with respect to the elements of the associated low-dimensional code, $\textbf{z}$, have a (complicated) analytical solution that can be calculated by THEANO using auto-differentiation. In other words, the sensitivity (Jacobian) matrix $\displaystyle\frac{d\textbf{X}}{d\textbf{z}}$ is immediately available. This may prove useful for gradient-based inversion using a forward solver equipped with an adjoint model.

In addition, it is worth noting that using a deep generative model to compress high dimensional quantities into a low-dimensional standard normal (or uniform) base is not only useful for inversion purposes but also holds promise for the so-called prediction-focused approach \citep[e.g.,][]{Satija2017}. The latter seeks to build a statistical model that links past data variables and prediction data variables, thereby enabling to make predictions without inverting for material properties. Lastly, note that our approach is not limited to the MPS framework but could potentially be applied to any prior with discrete structure.

\section{Conclusion}
\label{conclusion}

This paper presents a novel parametric low-dimensional representation of complex binary geologic media that can be used for fast sampling of complex geostatistical models, which is necessary for efficient probabilistic inversion. A deep neural network of the variational autoencoder type is used to define a low-dimensional manifold that is restricted to model realizations that agree well with the training set. By comparing other parametric dimensionality reduction (DR) techniques for unconditional geostatistical simulation of a channelized prior model, we find that our representation is superior to the PCA, optimization-PCA (OPCA) and discrete cosine transform (DCT). Since constructing the parameterization requires a training set of several tens of thousands of model realizations obtained by multiple point statistics simulation, our DR approach is specifically designed for probabilistic (or deterministic) inversion rather than for unconditional (or point-conditioned) geostatistical simulation. Synthetic inversion case studies involving 2D steady flow and 3D transient hydraulic tomography are used to illustrate the effectiveness of the proposed DR-based probabilistic inversion approach. For the 2D case study, our new approach provides better results than current state-of-the-art inversion by sequential geostatistical resampling (SGR). Inversion results for the 3D application are also encouraging. Future work will focus on improvements such as alleviating the need of using a large training model set to build the parameterization, extension to multi-categorical and continuous variables and defining a low-dimensional representation that is local rather than global. We expect that a local representation in which a given low-dimensional variable only influences a sub-region of the generated model realization will provide a much improved inversion performance.

\acknowledgments
Python codes of the proposed dimensionality reduction (DR) and DR-based inversion approaches are available from the first author (and can be downloaded from \url{https://github.com/elaloy}). This work was partially supported by Agence Nationale de la Recherche through the grant ANR-16-CE23-0006 Deep in France. The third author is a Research Associate with the Belgian F.R.S.-F.N.R.S. We thank Mike Swarbrick Jones for sharing his codes (\url{https://github.com/mikesj-public}), one of which served as a starting point for our devised VAECNN. We are grateful to Hai Xuan Vo and Louis Durlofsky for sharing their OPCA code, and to Xue Li and her co-authors for sharing their GC MPS algorithm. A temporary academic license of the DeeSse MPS code can be obtained upon request to one of its developers (Gr\'egoire Mariethoz, Philippe Renard, Julien Straubhaar). We thank Laurent Lemmens for sharing his MPH and CF calculation routines. Finally, we would like to thank the two anonymous referees for their useful comments.
%
%
%
%



%

{}

\end{document}